\pdfoutput=1

\documentclass[11pt]{article}

\usepackage[preprint]{acl}

\usepackage[utf8]{inputenc}        
\usepackage[T1]{fontenc}           

\usepackage{times}                 
\usepackage{latexsym}              
\usepackage{inconsolata}           

\usepackage{amsmath, amssymb}      
\usepackage{amsfonts}              
\usepackage{bm}

\usepackage{booktabs}              
\usepackage{multirow}             
\usepackage{array}                
\usepackage{siunitx}               

\usepackage{graphicx}              
\usepackage{caption}               
\usepackage{float}
\usepackage{wrapfig}
\usepackage{subfigure}
\usepackage{subcaption}

\usepackage{algpseudocode}

\usepackage{enumitem}              
\usepackage{dirtytalk}             

\usepackage{nicefrac}             
\usepackage{microtype}            

\usepackage{hyperref}               
\usepackage{url}                    

\usepackage{booktabs} 
\usepackage{graphicx} 
\usepackage{caption} 
\usepackage{tabularx} 
\usepackage{array}   

\usepackage{tikz}
\usetikzlibrary{positioning, arrows.meta, calc, shapes.geometric, shadows, backgrounds, fit}
\usepackage{fontawesome5}

\definecolor{paultolblue}{HTML}{332288}
\definecolor{paultolcyan}{HTML}{88CCEE}
\definecolor{paultolgreen}{HTML}{44AA99}
\definecolor{paultolorange}{HTML}{EE7733}
\definecolor{paultolpurple}{HTML}{AA4499}
\definecolor{paultolred}{HTML}{CC6677}
\definecolor{darktext}{gray}{0.2}

\colorlet{paultolblue_plate}{paultolblue!15!white}
\colorlet{paultolgreen_plate}{paultolgreen!15!white}
\colorlet{paultolorange_plate}{paultolorange!20!white}
\colorlet{paultolpurple_plate}{paultolpurple!15!white}
\colorlet{paultolred_plate}{paultolred!15!white}

\colorlet{paultolblue_nodefill}{paultolblue!45!white}
\colorlet{paultolgreen_nodefill}{paultolgreen!45!white}
\colorlet{paultolorange_nodefill}{paultolorange!55!white}
\colorlet{paultolpurple_nodefill}{paultolpurple!45!white}
\colorlet{paultolred_nodefill}{paultolred!45!white}

\tikzset{
  node_base/.style={
      draw=gray!70,
      rounded corners=3pt,
      inner sep=4pt, 
      minimum height=2.4em,
      font=\scriptsize,
      align=center,
      text=darktext,
      drop shadow={opacity=0.15, shadow xshift=0.15mm, shadow yshift=-0.15mm, fill=gray!30}
  },
  stage_main_node/.style={ 
      node_base,
      text width=5cm 
  },
  io_node/.style={
      stage_main_node, fill=paultolgreen_nodefill, text width=4.5cm
  },
  preprocess_node/.style={
      stage_main_node, fill=paultolgreen_nodefill, text width=4.5cm
  },
  projection_node/.style={
      stage_main_node, fill=paultolblue_nodefill
  },
  core_lstm_node/.style={
      stage_main_node, fill=paultolblue_nodefill, font=\scriptsize\bfseries,
  },
  parallel_block_node/.style={
      node_base,
      text width=2.6cm, 
      minimum height=3.2em 
  },
  max_pool_node/.style={parallel_block_node, fill=paultolorange_nodefill},
  attention_node/.style={parallel_block_node, fill=paultolorange_nodefill},
  avg_pool_node/.style={parallel_block_node, fill=paultolorange_nodefill},
  combine_node/.style={
      stage_main_node, fill=paultolpurple_nodefill, text width=3.5cm
  },
  classifier_node/.style={
      stage_main_node, fill=paultolred_nodefill, text width=4.5cm
  },
  conn/.style={
      -{Stealth[scale=1.2]}, thick, draw=gray!80
  },
  dimtag/.style={
      font=\tiny\itshape, text=gray!80, midway, sloped,
      above, yshift=0.5pt, fill=white, inner sep=0.4pt, opacity=0.6, text opacity=1
  },
  dimtag_right/.style={dimtag, right, xshift=2pt, sloped=false, text width=1.5cm, align=center},
  stageplate/.style={
      fill=#1, rounded corners=4pt, draw=gray!30, line width=0.3pt,
      inner sep=4mm 
  }
}

\title{Guided Perturbation Sensitivity (GPS): Detecting Adversarial Text via Embedding Stability and Word Importance}

\author{Bryan E. Tuck \\
  University of Houston  \\
  Houston, TX, USA \\
  \texttt{betuck@uh.edu} \\\And
  Rakesh M. Verma \\
  University of Houston \\
  Houston, TX, USA \\
  \texttt{rverma@uh.edu} \\}

\begin{document}

\maketitle
\footnotetext[0]{\textbf{Accepted to AAAI 2026.} Code: \url{https://github.com/ReDASers/Guided-Perturbation-Sensitivity}.}

\begin{abstract}
Adversarial text attacks remain a persistent threat to transformer models, yet existing defenses are typically attack-specific or require costly model retraining, leaving a gap for attack-agnostic detection. We introduce Guided Perturbation Sensitivity (GPS), a detection framework that identifies adversarial examples by measuring how embedding representations change when important words are masked. GPS first ranks words using importance heuristics, then measures embedding sensitivity to masking top-$k$ critical words, and processes the resulting patterns with a BiLSTM detector. Experiments show that adversarially perturbed words exhibit disproportionately high masking sensitivity compared to naturally important words. Across three datasets, three attack types, and two victim models, GPS achieves over 85\% detection accuracy and demonstrates competitive performance compared to existing state-of-the-art methods, often at lower computational cost. Using Normalized Discounted Cumulative Gain (NDCG) to measure perturbation identification quality, we demonstrate that gradient-based ranking significantly outperforms attention, hybrid, and random selection approaches, with identification quality strongly correlating with detection performance for word-level attacks ($\rho = 0.65$). GPS generalizes to unseen datasets, attacks, and models without retraining, providing a practical solution for adversarial text detection.
\end{abstract}

\section{Introduction}
\label{sec:introduction}

\begin{figure}[ht]
  \centering
  \resizebox{0.78\columnwidth}{!}{%
  \begin{tikzpicture}[
      scale=0.95,
      important/.style={text=paultolred, font=\bfseries},
      box/.style={draw, rounded corners=1pt, inner sep=2pt, font=\footnotesize},
      compare_note/.style={font=\scriptsize, text=paultolgreen!60!black}
  ]
  
  \node[align=left, text width=5.5cm, font=\footnotesize] (benign_text) at (0,0.8) {
      Benign: The film was severely \textcolor{paultolgreen!70!black}{awful}.
  };
  
  \node[align=left, text width=5.5cm, font=\footnotesize] (adv_text) at (0,0) {
      Adversarial: The film was severely \textcolor{paultolred}{terrible}.
  };
  
  \draw[gray!70] (-2.8,-0.7) -- (2.8,-0.7); 
  
  \node[text width=5.2cm] (sensitivity_section) at (0,-2.0) { 
      \begin{minipage}{\linewidth}
      \raggedright
      \footnotesize
      1. Masked Sensitivity (Top-K from Adv. Text):\\
      \vspace{1pt}
      \begin{tabular}{@{}ll@{}}
      "film" & 0.010 \\
      "severely" & 0.012 \\
      "\textcolor{paultolred}{terrible}" & \textcolor{paultolred}{0.028} (High)\\[3pt]
      \multicolumn{2}{l}{\hspace{1em}[\textit{benign} \textcolor{paultolgreen!70!black}{"awful"}: 0.014]}
      \end{tabular}
      \end{minipage}
  };
  
  \draw[gray!70] (-2.8,-3.3) -- (2.8,-3.3); 
  
  \node[align=center, font=\footnotesize] (trace_title) at (0,-3.7) {2. Feature Trace from Adversarial (Top-K):}; 
  
  \begin{scope}[shift={(0,-4.4)}] 
      \foreach \x/\word/\imp/\sens/\color in {
          -2.0/film/0.08/0.010/paultolblue,
          -0.0/severely/0.10/0.012/paultolblue, 
           2.0/terrible/0.32/0.028/paultolred%
      } {
          \draw[fill=\color!25] (\x-0.4,0) rectangle +(0.35, \imp*1.5);
          \draw[fill=\color!55] (\x+0.05,0) rectangle +(0.35, \sens*10);
          \node[font=\tiny, align=center] at (\x, -0.25) {\word};
          \node[font=\tiny, align=center, rotate=90] at (\x-0.55, \imp*0.75) {\imp};
          \node[font=\tiny, align=center, rotate=90] at (\x+0.55, \sens*5) {\sens};
      }
      \node[font=\tiny, text=gray] at (0, -0.55) {(Importance | Sensitivity)};
  \end{scope}
  
  \draw[rounded corners, fill=paultolpurple!15] (-2.8,-5.6) rectangle (2.8,-5.1); 
  \node[font=\footnotesize] at (0,-5.35) {3. BiLSTM (processes adversarial trace)}; 
  
  \draw[thick, ->] (0,-5.6) -- (0,-5.9); 
  \node[box, fill=paultolred!15, text=darktext] at (0,-6.2) {Prediction: Adversarial}; 
  
  \end{tikzpicture}
  }
  \captionsetup{skip=6pt, font=small, labelfont=bf}
  \caption{Guided Perturbation Sensitivity (GPS) workflow for detecting adversarial text using top-$3$. After identifying important words (not shown), GPS measures embedding sensitivity to masking (1). The adversarial word \textit{terrible} shows significantly higher sensitivity than its benign counterpart \textit{awful}. The feature trace (2) displays paired bars for each word; the left bar shows importance scores, and the right bar shows sensitivity values, revealing distinctive patterns for adversarial words. A BiLSTM classifier (3) processes this trace to detect manipulated text.}

  \label{fig:intro_overview}
\end{figure}
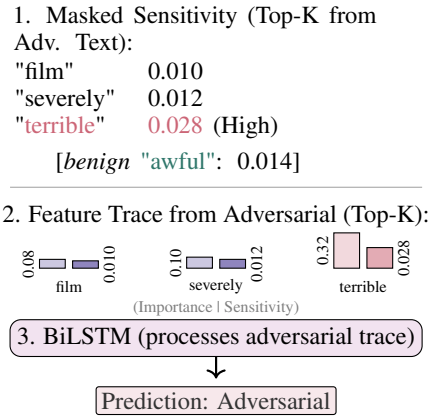

A single word substitution can fool a state-of-the-art transformer into classifying a positive movie review as negative, or trick a spam filter into allowing malicious content through. While transformer models achieve remarkable performance on NLP benchmarks, they remain surprisingly brittle to \emph{adversarial examples}—subtle, meaning-preserving perturbations that flip predictions \cite{Goodfellow2014ExplainingAH}. As these models are deployed in high-stakes applications from healthcare diagnostics to financial fraud detection, such vulnerabilities pose serious risks where even occasional failures can erode trust, enable manipulation, or trigger costly security incidents~\cite{tuck_adv_mindset2025}.

The fundamental challenge in adversarial text detection lies in distinguishing malicious edits from natural language variation. Unlike vision, where perturbations are continuous pixel modifications, text attacks operate in discrete lexical space, requiring manipulations that preserve semantics while remaining imperceptible to human readers \cite{Morris2020TextAttackAF}. Word substitutions like changing \say{excellent} to \say{great} or character-level edits like \say{moive} for \say{movie} can completely flip model predictions while appearing innocuous.

Current detection approaches face a critical limitation: they either assume knowledge of specific attack patterns \cite{Jones2020, Wang2021SynEnc} or require expensive model retraining \cite{Ye2020, Zeng2023}. Methods that analyze output-layer signals often overfit to particular attacks and fail to generalize \cite{Mosca2022ThatIA}, while gradient-based detectors overlook the rich sequential structure of adversarial manipulations \cite{shen2023textshield}. We need a detection approach that exploits the fundamental instability of adversarial examples without requiring attack-specific knowledge.

We build on a crucial theoretical foundation: adversarial examples reside near decision boundaries in regions of high curvature, where small perturbations cause dramatic classification changes \cite{Fawzi2018EmpiricalSO, Bell2024PersistentCU}. We hypothesize that this instability extends beyond decision boundaries into the representation space itself. By strategically masking important words, adversarial examples should exhibit disproportionate sensitivity compared to naturally important words in benign text, revealing their artificial nature through instability patterns.

This insight motivates \textbf{Guided Perturbation Sensitivity (GPS)}~(Figure \ref{fig:intro_overview}): a detection framework that identifies adversarial examples by measuring how embedding representations change when important words are masked. GPS first ranks words using importance-based methods, then measures embedding sensitivity to masking top-$k$ critical words, and processes these sensitivity patterns with a BiLSTM detector. GPS detects adversarial examples without requiring knowledge of specific attack models or model retraining, generalizing across attacks, datasets, and models.

\textbf{Our contributions are as follows:}
\begin{itemize}
    \item We introduce \textbf{Guided Perturbation Sensitivity (GPS)} (\S\ref{sec:methodology}), a detection method that identifies adversarial examples by measuring embedding instability under targeted word masking, requiring no modification of the target model.
    
    \item We provide empirical evidence that \textbf{adversarial examples exhibit approximately 2× higher embedding sensitivity} (\S\ref{sec:sens_analys}) to strategic word masking compared to benign inputs, empirically linking decision boundary theory to the representation level.
    
    \item Through \textbf{comprehensive evaluation across 18 experimental configurations} (\S\ref{sec:setup}--\S\ref{sec:comp_tradeoff}), we demonstrate that GPS achieves 85\%+ detection accuracy across three datasets, three attack types, and two models, with superior generalization and computational efficiency reaching 98\% performance at just $K=5$ words.
    
    \item We reveal \textbf{fundamental differences in detection mechanisms} (\S\ref{sec:ndcg_analysis}, \S\ref{sec:corr_analysis}) showing gradient-based importance ranking achieves strong correlation ($\rho > 0.65$) between perturbation identification and detection performance for word-level attacks, while character-level attacks require different strategies.
\end{itemize}

\section{Related Work}
\label{sec:rw}

\paragraph{Adversarial vulnerability.}
Adversarial machine learning emerged in \citet{Huang2011AdvML} and gained prominence in computer vision~\cite{Szegedy2014}, attributed to neural network linearity~\cite{Goodfellow2014ExplainingAH}. This led to optimization-based attacks like Carlini-Wagner~\cite{Carlini2017}. While these concepts extend to sequential data~\cite{Papernot2016}, text requires semantic and grammatical preservation during perturbation.

\paragraph{Adversarial Text Attacks.}
Text attacks operate at character, word, and sentence level. Character methods include gradient-guided flips~\cite{Ebrahimi2018}, importance-based edits~\cite{Gao2018}, and Charmer~\cite{rocamora2024revisiting}, which achieves high success rates against both BERT and LLMs. Word-level approaches evolved from genetic algorithms~\cite{Alzantot2018} to importance-ranking systems~\cite{Jin2020} and contextual substitutions~\cite{li-etal-2020-bert-attack}. Recent advances include GBDA~\cite{guo-etal-2021-gradient}, optimizing distributions of adversarial examples, and ATGSL~\cite{li-etal-2023-adversarial}, which balances attack effectiveness with text quality using simulated annealing and fine-tuned language models. These attacks are still effective with success rates often exceeding 90\% against state-of-the-art transformers while maintaining semantic preservation, making them particularly challenging targets for detection systems~\cite{10.1145/3579987.3586567}.

\paragraph{Defense strategies against adversarial attacks.}
Existing NLP defenses fall into three camps: adversarial training~\cite{Miyato2017}, certified robustness~\cite{Jia2019,Ye2020,Zhang2024}, and post-hoc detection. Detection approaches range from surface statistics like word frequency~\cite{Mozes2021} and logit irregularities~\cite{Mosca2022ThatIA} to attribution signals~\cite{10.1145/3729235} and ensemble methods combining multiple gradient-based importance measures like TextShield~\cite{shen2023textshield}. Recent work explores loss landscape geometry:~\cite{zheng-etal-2023-detecting} measures sharpness by maximizing local loss increments, while TextDefense~\cite{textdefense} uses dispersion of word-importance scores to flag suspicious inputs. These detectors either assume attack-specific artifacts, require additional optimization loops, or treat model outputs as static signals; none directly measure how the model's internal representations respond to targeted input modifications. GPS addresses this gap by probing dynamic embedding responses to guided masking, cleanly separating word-level from character-level behaviors and offering a scalable alternative to sharpness and ensemble-based methods.

\section{Methodology}
\label{sec:methodology}

Our methodology detects adversarial text by analyzing how targeted word perturbations affect transformer embedding stability. We hypothesize that adversarially manipulated words exhibit unusual importance patterns and cause disproportionate embedding shifts compared to naturally important words. GPS identifies influential words using importance heuristics, measures embedding stability by sequentially masking top-ranked words, and processes the resulting sensitivity patterns with a detector. Our evaluation of gradient, attention, hybrid, and random selection reveals that gradient-based methods outperform attention-based approaches for word-level attacks.

\subsection{Reference Embeddings}
Given an input text $\mathcal{T}= (w_{1},\dots,w_{N})$ (either benign or potentially adversarial) and a frozen transformer model $f$, we first compute its reference sentence embedding. This is obtained by averaging the final hidden states of its non-special subtokens:
\begin{equation}
    \mathbf{e}(\mathcal{T}) = \frac{1}{|\Omega|}\sum_{i\in\Omega}\mathbf{h}_{i}^{(L)} \in\mathbb{R}^{d},
\end{equation}
where $\mathbf{h}_{i}^{(L)}$ is the final layer hidden state for the $i$-th subtoken, $\Omega = \{i \mid \text{subtoken}~i~\text{is not a special token}\}$, and $d$ is the embedding dimension. For adversarial detection tasks, we compute reference embeddings $\mathbf{e}_{\text{ben}}$ and $\mathbf{e}_{\text{adv}}$ for the benign and adversarial versions, respectively.

\subsection{Identifying Influential Words with Importance Heuristics}
\label{sec:importance}

To focus our sensitivity analysis on the most relevant words, we first rank all words $w_k$ within the text $\mathcal{T}$ by their predicted importance to the model's decision. We compute an importance score $\alpha_k$ for each word using one of four post-hoc heuristics that require no model modification. The choice of heuristic significantly impacts detection performance, as it determines which words undergo sensitivity testing.

We evaluate four importance ranking strategies to identify the most effective approach for adversarial detection:

\paragraph{Gradient Attribution.} Based on the intuition that words critical to the model's prediction exhibit large gradients, we compute importance scores following \citet{Simonyan2013DeepIC}. We backpropagate the gradient of the cross-entropy loss $\ell(\mathcal{T})$ with respect to input embeddings $\mathbf{e}_j$ while keeping model $f$ frozen. The importance score for word $w_k$ sums the $\ell_2$-norms of gradients across its constituent subtokens $j \in \mathcal{S}_k$:
\begin{equation}
    \alpha_k^{\text{sal}} = \sum_{j\in\mathcal{S}_k}\bigl\|\nabla_{\mathbf{e}_j}\ell(\mathcal{T})\bigr\|_2,
    \label{eq:saliency}
\end{equation}
where $\ell(\mathcal{T})$ is the cross-entropy loss with respect to the predicted class, and $\mathcal{S}_k$ represents subtokens comprising word $w_k$. We sum over subtokens so that morphologically complex words contribute proportionally to the importance signal. This requires white-box access; surrogate-based saliency can substitute in black-box settings without modifying the detector. Our experiments demonstrate this approach most effectively identifies adversarially perturbed words.

\paragraph{Attention Rollout.} To capture information flow through transformer layers, we employ attention rollout \citep{abnar-zuidema-2020-quantifying}, which aggregates attention patterns across layers to estimate overall token attention. For each layer $\ell$, we compute the head-averaged attention matrix $\mathbf{A}_{\text{avg}}^{(\ell)}$, incorporate residual connections, and row-normalize: $\hat{\mathbf{A}}^{(\ell)} = \text{row-normalize}(0.5 \cdot \mathbf{A}_{\text{avg}}^{(\ell)} + 0.5 \cdot \mathbf{I})$. These matrices are recursively multiplied across layers: $\mathbf{R} = \hat{\mathbf{A}}^{(1)} \times \cdots \times \hat{\mathbf{A}}^{(L)}$. The attention mass flowing to token $j$ is $a_j = \sum_i R_{ij}$, yielding word scores:
\begin{equation}
   \alpha_k^{\text{roll}} = \sum_{j \in \mathcal{S}_k} a_j.
   \label{eq:rollout}
\end{equation}

\paragraph{Grad-SAM.} Inspired by Grad-CAM \citep{gradcam} for vision, Grad-SAM \citep{gradsam} combines gradient information with attention weights to highlight tokens that are both attended to and important for the prediction. We capture attention weights $\mathbf{A}^{(\ell)}$ and their gradients $\nabla \mathbf{A}^{(\ell)}$ for each layer $\ell$ (obtained by backpropagating the predicted logit). We compute the element-wise product $\mathbf{G}^{(\ell)} = \nabla \mathbf{A}^{(\ell)} \odot \mathbf{A}^{(\ell)}$ for each layer. We aggregate these layer-wise products by averaging across all $L$ transformer layers, followed by averaging over all $H$ attention heads in the model. The resulting scores are finally summed for the subtokens $j \in \mathcal{S}_k$ corresponding to word $w_k$:
\begin{equation}
\alpha_k^{\text{gatt}}= \sum_{j\in\mathcal{S}_k} \sum_i \left(\frac{1}{H}\sum_{h=1}^{H}\frac{1}{L}\sum_{\ell=1}^{L}(\mathbf{G}_{h}^{(\ell)})_{ij}\right).
\label{eq:gradattn}
\end{equation}

\paragraph{Random Baseline.} As a control, we randomly select $K$ distinct words and assign them importance score $\alpha_k^{\text{rand}}=1$, with remaining words receiving $\alpha_k=0$. This baseline helps isolate the contribution of targeted word selection versus random masking.

\subsection{Sequential Sensitivity Profiling via Masking}
\label{sec:sensitivity_profiling}

After ranking words by importance $\boldsymbol{\alpha}$, we select the top $K$ most important words, denoted $\mathcal{I}_K=\text{TOP-}K(\boldsymbol{\alpha})$. Through ablation studies across $K$ values ranging from 5 to 50, we find that performance remains relatively stable across this range, with minimal degradation between $K=5$ and $K=50$ (\S~\ref{sec:comp_tradeoff}). We select $K=20$ for all experiments as it provides optimal accuracy while maintaining reasonable computational efficiency.

We then probe embedding stability with respect to these influential words through sequential masking. For each selected word index $k \in \mathcal{I}_K$, we create a masked version by replacing word $w_k$ with the model's [MASK] token, yielding embedding $\tilde{\mathbf{e}}_k = \mathbf{e}(\mathcal{T} \text{ with } w_k \text{ masked})$. The sensitivity $s_k$ quantifies the embedding space change caused by masking, measured as cosine distance between the reference embedding $\mathbf{e}(\mathcal{T})$ and masked embedding $\tilde{\mathbf{e}}_k$:

\begin{equation}
    s_k = 1 - \frac{\mathbf{e}(\mathcal{T})\cdot\tilde{\mathbf{e}}_k}
         {\|\mathbf{e}(\mathcal{T})\|_{2}\,
          \|\tilde{\mathbf{e}}_k\|_{2}}.
    \label{eq:sensitivity}
\end{equation}

Since we mask words individually, $s_k$ captures the specific impact of each word $w_k$ on the overall representation. We find that adversarially perturbed words exhibit disproportionately high sensitivity values compared to naturally important words, as they represent artificial manipulations that create unstable embedding regions.

\subsection{GPS Feature Tensor for Detection}
\label{sec:feature_tensor}

The sensitivity profiling yields a sensitivity score $s_k$ for each word $w_k$ among the top $K$ most important words. We combine these sensitivity scores with the corresponding importance scores $\alpha_k$ while preserving the original word order of the text $\mathcal{T}$. This results in two aligned sequences of length $N$ (the original number of words):

\begin{itemize}
\item The sensitivity sequence $\mathbf{s}=(s_{1},\dots,s_{N})$, where $s_k$ is the computed sensitivity if $k \in \mathcal{I}_K$, and $s_k=0$ otherwise.
\item The importance sequence $\boldsymbol{\alpha}=(\alpha_{1},\dots,\alpha_{N})$, where $\alpha_k$ is the computed importance score if $k \in \mathcal{I}_K$, and $\alpha_k=0$ otherwise.
\end{itemize}

We stack these two sequences column-wise to form an $N \times 2$ feature tensor $\mathbf{Z}=[\mathbf{s}\,\|\,\boldsymbol{\alpha}]$. This tensor retains the original positional information of each word while highlighting the sensitivity and importance of the words identified as most influential by the chosen heuristic. The resulting GPS features $\mathbf{Z}$ can serve as input to any classifier, from simple linear models to neural architectures.

\section{Sensitivity Analysis Results}
\label{sec:sens_analys}

\begin{table}[b]
\centering
\small
\resizebox{\columnwidth}{!}{
\begin{tabular}{lccc}
\hline
\textbf{Importance Method} & \textbf{Benign Mean} & \textbf{Adversarial Mean} & \textbf{Ratio} \\
\hline
Gradients & 0.014 & 0.028 & 1.932 \\
Attention Rollout & 0.014 & 0.028 & 1.912 \\
Grad-SAM & 0.014 & 0.027 & 1.836 \\
Random & 0.013 & 0.026 & 1.880 \\
\hline
\end{tabular}
}
\caption{Mean sensitivity values $s_k$ across importance methods with $K{=}20$. We compute sensitivity $s_k$ as cosine distance between original and masked embeddings (Eq.~\ref{eq:sensitivity}), then take the mean across all masked positions. Columns show averages across experiments (18 per method). Ratio is the mean of per-experiment ratios (adversarial/benign for each experiment). Results are averaged across 3 datasets, 3 attacks, and 2 models.}
\label{tab:gps_mean}
\end{table}

GPS tests whether adversarial examples exhibit measurably different embedding stability than benign text. Table \ref{tab:gps_mean} validates this: adversarial examples demonstrate 1.89× higher sensitivity on average, with 88.9\% of experiments showing increased instability. Instability ratios cluster tightly across methods (1.836--1.932). Notably, random word selection achieves comparable performance (1.880×) to gradient-based and attention-based methods. This empirically extends established decision boundary instability from classification to representation space. Embedding instability is an intrinsic property of adversarial examples rather than dependent on any particular importance heuristic, which lets GPS achieve reliable detection.

\begin{table}[t]
\centering
\small
\resizebox{\columnwidth}{!}{%
\begin{tabular}{@{}p{0.2\columnwidth}p{0.75\columnwidth}@{}}
\toprule
\textbf{Component} & \textbf{Options} \\
\midrule
Datasets & IMDB \cite{imdb} (binary sentiment) \\
& AG News (4-way topic) \cite{Zhang2015CharCNN} \\
& Yelp Polarity (binary review) \cite{Zhang2015CharCNN} \\
\midrule
Attacks & TextFooler \cite{Jin2020} (word substitution) \\
& BERT-Attack \cite{li-etal-2020-bert-attack} (contextual) \\
& DeepWordBug \cite{Gao2018} (char-level) \\
\midrule
Models & RoBERTa-base \cite{Liu2019RoBERTaAR} \\
& DeBERTa-V3-base \cite{He2021DeBERTaV3ID} \\
\midrule
Importance & Gradient Attribution \cite{Simonyan2013DeepIC} \\
Heuristics & Attention-Rollout \cite{abnar-zuidema-2020-quantifying} \\
& Grad-SAM \cite{gradsam} \\
& Random selection \\
\midrule
Baselines & TextShield \cite{shen2023textshield} \\
& Sharpness-based detection \cite{zheng-etal-2023-detecting} \\
\bottomrule
\end{tabular}%
}
\caption{Experimental matrix. For data, attack, and model configuration, we generate 5,000 balanced training (20\% held out for validation) and 1,000 test samples. Our generated adversarial examples exclusively comprise true adversarial samples that successfully deceived the target model; failed perturbation attempts that did not achieve misclassification are excluded from the corpus.}
\label{tab:experimental_components}
\end{table}

\section{Experimental Setup}
\label{sec:setup}

We evaluate GPS across the comprehensive experimental matrix shown in Table~\ref{tab:experimental_components}, designed to assess robustness across diverse adversarial scenarios while controlling for architectural, linguistic, and attack-specific variations. Our model selection strategy targets generalization across different transformer architectures: we leverage architectural differences between the models, with DeBERTa's disentangled attention mechanisms separating content and position information and its larger parameter count providing a contrast to standard attention and smaller capacity used in RoBERTa.

For adversarial sample generation, we employ TextAttack~\cite{Morris2020TextAttackAF} across most attack methods, with the exception of BERT-Attack.\footnote{BERT-Attack's search over up to $K\!=\!48$ substitutions per sub-word can explode combinatorially (e.g., $48^{4}$ candidates for a four-piece token), driving runtimes prohibitively high. We therefore use the TextDefender~\cite{Li2021SearchingFA} implementation, which employs word-level swaps to maintain computational feasibility.} Our importance heuristic evaluation spans gradient-based, attention-based, and hybrid approaches, with random selection serving as a lower-bound control to validate that GPS performance stems from meaningful semantic signal rather than dataset artifacts.

For baselines, we utilize state-of-the-art adversarial detection methods with proven superiority over multiple contemporary approaches. TextShield and sharpness-based detection have demonstrated effectiveness against 8+ established methods, including MD~\cite{NEURIPS2018_abdeb6f5}, DISP~\cite{zhou-etal-2019-learning}, FGWS~\cite{Mozes2021}, and WDR~\cite{Mosca2022ThatIA}, providing strong baselines for GPS evaluation.

\section{Detection Performance}
\label{sec:detection_performance}

Having established that adversarial examples exhibit embedding instability, we now demonstrate that this translates into practical detection performance (Table~\ref{tab:accuracy_results}). We evaluate GPS using a BiLSTM model to classify sensitivity-importance traces $\mathbf{Z} \in \mathbb{R}^{N\times2}$ as benign or adversarial. BiLSTM efficiently captures sequential dependencies while handling variable-length texts and maintaining low parameter counts (257,154 parameters); we train using AdamW optimizer (lr=$5 \times 10^{-4}$), batch size 32, with 10\% of training data reserved for validation, and early stopping on validation F1-score with 5-epoch patience over a maximum of 40 epochs. We derive traces from our four importance heuristics with $K{=}20$ words and compare against TextShield~\cite{shen2023textshield}, an ensemble of four LSTMs processing gradient-based features,\footnote{As official code was unavailable, we reimplemented TextShield following the original paper's specifications.} and Sharp~\cite{zheng-etal-2023-detecting}, a sharpness-based detector measuring local loss landscape curvature.

Gradient-based heuristics consistently outperform attention-based methods, echoing critiques of attention as a direct proxy for predictive importance~\cite{Jain2019AttentionIN}. Gradient Attribution and Grad-SAM match or exceed state-of-the-art baselines, confirming that embedding instability is most evident when perturbing words critical to model predictions. GPS's superior performance over TextShield's ensemble approach and Sharp's loss landscape analysis validates that targeted embedding perturbations provide reliable adversarial detection signals. Contrasting behaviors reveal key insights: gradient-based GPS consistently surpasses attention-based approaches on semantic substitution attacks, while attention-guided GPS remains competitive against character-level DeepWordBug attacks on IMDB. TextShield and Sharp experience significant performance degradation on IMDB's character-level attacks, performing worse than random selection, yet maintain stronger performance on AG News and Yelp. Effective adversarial detection requires aligning detection mechanisms with both specific embedding disruptions and dataset characteristics.

\begin{table}[t]
    \centering
    \small
    \resizebox{\columnwidth}{!}{%
    \begin{tabular}{llcrrrr|rr}
    \toprule
    \textbf{Dataset} & \textbf{Model} & \textbf{Attack} & \textbf{Rand} & \textbf{Attn} & \textbf{GS} & \textbf{Grad} & \textbf{TS} & \textbf{Sharp} \\
    \midrule
    \multirow{6}{*}{\rotatebox{90}{\textbf{AG News}}} & \multirow{3}{*}{RoBERTa} & BA & 0.717 & 0.714 & 0.788 & 0.845 & \textbf{0.846} & 0.837 \\
     &  & TF & 0.775 & 0.796 & 0.843 & 0.887 & \textbf{0.893} & 0.874 \\
     &  & DWB & 0.781 & 0.772 & 0.864 & \textbf{0.895} & 0.883 & 0.860 \\
     \cmidrule(lr){2-9}
     & \multirow{3}{*}{DeBERTa} & BA & 0.729 & 0.744 & 0.801 & 0.839 & \textbf{0.840} & 0.786 \\
     &  & TF & 0.798 & 0.804 & 0.821 & \textbf{0.884} & 0.883 & 0.832 \\
     &  & DWB & 0.782 & \textbf{0.902} & 0.860 & 0.897 & 0.878 & 0.812 \\
     \midrule
     \multirow{6}{*}{\rotatebox{90}{\textbf{IMDB}}} & \multirow{3}{*}{RoBERTa} & BA & 0.741 & 0.755 & 0.830 & 0.845 & 0.783 & \textbf{0.873} \\
     &  & TF & 0.846 & 0.846 & 0.913 & \textbf{0.919} & 0.870 & 0.888 \\
     &  & DWB & 0.936 & 0.935 & \textbf{0.959} & 0.958 & 0.813 & 0.859 \\
     \cmidrule(lr){2-9}
     & \multirow{3}{*}{DeBERTa} & BA & 0.606 & 0.621 & 0.698 & 0.755 & 0.731 & \textbf{0.797} \\
     &  & TF & 0.731 & 0.757 & 0.803 & \textbf{0.859} & 0.756 & 0.800 \\
     &  & DWB & 0.929 & 0.930 & 0.938 & \textbf{0.968} & 0.775 & 0.775 \\
     \midrule
     \multirow{6}{*}{\rotatebox{90}{\textbf{Yelp}}} & \multirow{3}{*}{RoBERTa} & BA & 0.694 & 0.714 & 0.812 & 0.836 & 0.832 & \textbf{0.910} \\
     &  & TF & 0.774 & 0.783 & 0.860 & 0.899 & 0.849 & \textbf{0.912} \\
     &  & DWB & 0.781 & 0.771 & 0.878 & \textbf{0.927} & 0.874 & 0.895 \\
     \cmidrule(lr){2-9}
     & \multirow{3}{*}{DeBERTa} & BA & 0.772 & 0.804 & 0.844 & 0.870 & 0.826 & \textbf{0.905} \\
     &  & TF & 0.771 & 0.773 & 0.865 & \textbf{0.917} & \textbf{0.917} & 0.911 \\
     &  & DWB & 0.793 & 0.815 & 0.856 & \textbf{0.931} & 0.902 & 0.893 \\
    \bottomrule
    \end{tabular}%
    }
    \caption{
    Accuracy across datasets, models, attacks, and strategies with $K{=}20$.
    Best accuracy per condition is shown in \textbf{bold}.
    Attacks: BA (BERT-Attack), DWB (DeepWordBug), TF (TextFooler).
    Our Strategies: Rand (Random), Attn (Attention), GS (Grad-SAM),
    Grad (Gradient). Baselines: TS (TextShield), Sharp (Sharpness-based).
    }
    \label{tab:accuracy_results}
    \end{table}

\begin{table}[t]
\centering
\small
\resizebox{\columnwidth}{!}{%
\begin{tabular}{l cc cc cc}
\toprule
 & \multicolumn{2}{c}{\textbf{GPS (Grad)}} & \multicolumn{2}{c}{\textbf{TS}} & \multicolumn{2}{c}{\textbf{Sharp}} \\
\cmidrule(lr){2-3} \cmidrule(lr){4-5} \cmidrule(lr){6-7}
\textbf{Transfer Setting} & \textbf{In} & \textbf{Out} & \textbf{In} & \textbf{Out} & \textbf{In} & \textbf{Out} \\
\midrule
\multicolumn{7}{@{}l}{\textbf{R1: Dataset Shift}} \\
Yelp → IMDB & 0.883 & \textbf{0.878} & 0.838 & 0.806 & \textbf{0.913} & 0.755 \\
IMDB → Yelp & \textbf{0.902} & \textbf{0.855} & 0.822 & 0.848 & 0.886 & 0.765 \\
\midrule
\multicolumn{7}{@{}l}{\textbf{R2: Attack Shift}} \\
TF → DWB & \textbf{0.904} & \textbf{0.915} & 0.841 & 0.799 & 0.829 & 0.836 \\
DWB → TF & \textbf{0.943} & \textbf{0.875} & 0.831 & 0.816 & 0.835 & 0.829 \\
TF → BA & \textbf{0.904} & \textbf{0.839} & 0.841 & 0.794 & 0.829 & 0.839 \\
BA → TF & \textbf{0.844} & \textbf{0.875} & 0.807 & 0.855 & 0.839 & 0.829 \\
DWB → BA & \textbf{0.943} & 0.649 & 0.831 & 0.729 & 0.835 & \textbf{0.839} \\
BA → DWB & \textbf{0.844} & \textbf{0.862} & 0.807 & 0.815 & 0.839 & 0.836 \\
\midrule
\multicolumn{7}{@{}l}{\textbf{R3: Model Shift}} \\
RoBERTa → DeBERTa & \textbf{0.886} & \textbf{0.827} & 0.835 & 0.822 & 0.835 & 0.758 \\
DeBERTa → RoBERTa & \textbf{0.852} & \textbf{0.880} & 0.813 & 0.798 & 0.841 & 0.774 \\
\bottomrule
\end{tabular}%
}
\caption{Generalization performance of adversarial text detection methods across transfer settings with $K{=}20$. We evaluate three detection approaches: GPS (Ours, using Grad importance), TS (TextShield), and Sharp (sharpness-based detection). R1 tests cross-dataset generalization, R2 evaluates cross-attack generalization, and R3 examines cross-encoder transferability. In/Out columns show in-domain and out-of-domain F1 scores, respectively. BA (BERT-Attack), DWB (DeepWordBug), TF (TextFooler).}
\label{tab:generalization_results}
\end{table}

\begin{figure}[t]
    \centering
    \includegraphics[width=\columnwidth]{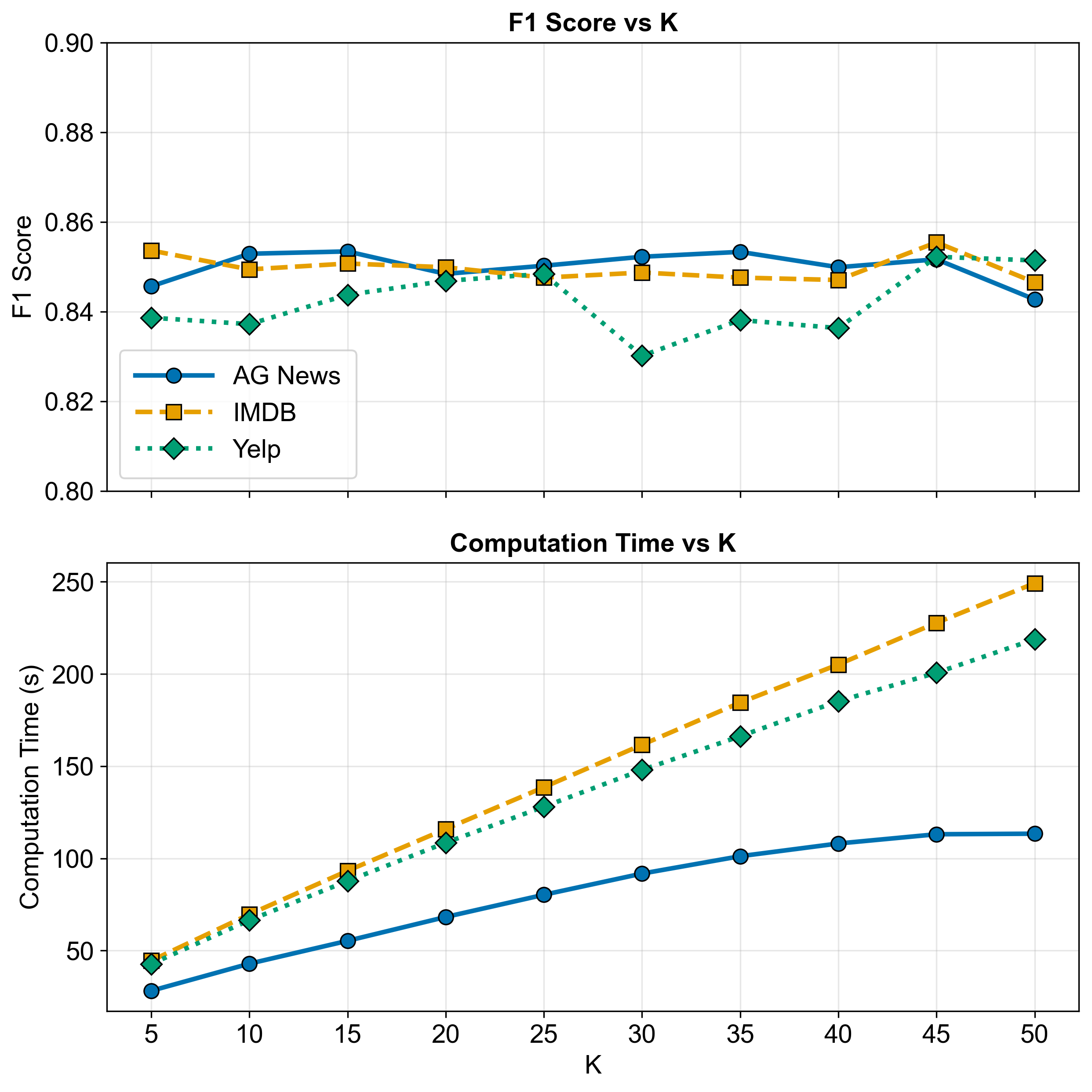} 
    \caption{Performance vs efficiency trade-off for GPS across different $K$ values on BERT-Attack adversarial examples using RoBERTa. The annotation box shows baseline computation times for comparison. GPS (28--249s) provides competitive timing with Sharp (51--86s) while significantly outperforming TextShield (111--233s), with the flexibility to trade computation time for detection accuracy via the $K$ parameter.}
    \label{fig:bertattack_comp}
\end{figure}

\subsection{Generalization Evaluation}
\label{sec:ablations}

Real-world deployment requires detectors that generalize beyond training conditions; if GPS only works on familiar datasets, attacks, or models, its practical utility is severely limited. We evaluate GPS's generalization capabilities across three dimensions using gradient attribution, our best-performing heuristic, and compare against TextShield and Sharp (Table~\ref{tab:generalization_results}). Dataset shifts (R1) involve training on adversarial examples from one dataset (combining TextFooler, DeepWordBug, and BERT-Attack on RoBERTa) and testing on another dataset. Attack shifts (R2) train detectors on one attack type across Yelp and IMDB datasets and test on a different attack type. Model shifts (R3) train on one transformer architecture and test on another, using all datasets and attacks.

GPS shows robust generalization across most transfer scenarios. Embedding sensitivity patterns reflect adversarial manipulation properties rather than dataset-specific artifacts, showing that gradient-based importance captures universal adversarial signatures. Cross-attack generalization shows GPS effectively transfers between word-level attacks (TextFooler, BERT-Attack), though performance drops significantly when transferring from character-level to contextualized semantic substitution attacks (DWB→BA). This reflects model-attack asymmetry: DeepWordBug is the weakest attack (Table~\ref{tab:accuracy_results}), and transfer from weaker to stronger attacks is inherently limited, consistent with stronger-to-weaker transfer observed. Cross-architecture results reveal an asymmetry also: positive transfer from DeBERTa to RoBERTa reflects DeBERTa's larger parameter capacity and disentangled attention mechanisms providing richer adversarial representations that generalize to smaller architectures. Sharp's fixed-threshold loss landscape method shows particular vulnerability to dataset and model shifts, maintaining stable performance only where loss sharpness ordering between benign and adversarial samples remains consistent.

\begin{figure*}[t]
    \centering
    \includegraphics[width=\textwidth]{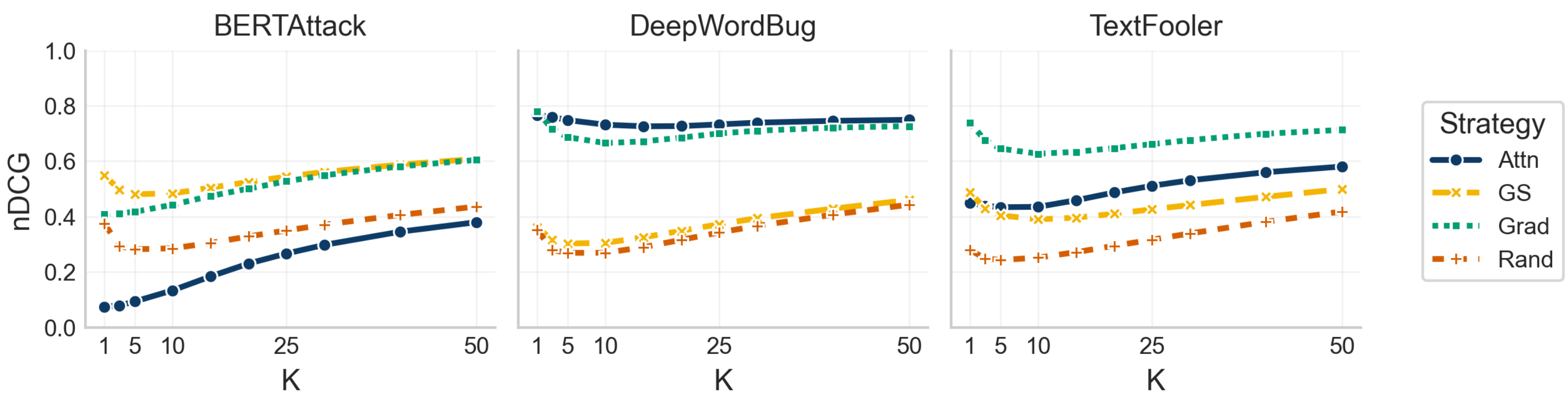}
    \caption{NDCG@k performance for ranking perturbed words on Yelp with RoBERTa across BERT-Attack, DeepWordBug, and TextFooler. Higher NDCG values indicate better ranking quality of truly perturbed words. Strategies: Rand (Random), Attn (Attention), GS (Grad-SAM), Grad (Gradient). Similar patterns hold across other dataset-model combinations.}
    \label{fig:ndcg_performance}
\end{figure*}

\begin{figure*}[t]
    \centering
    \includegraphics[width=\textwidth]{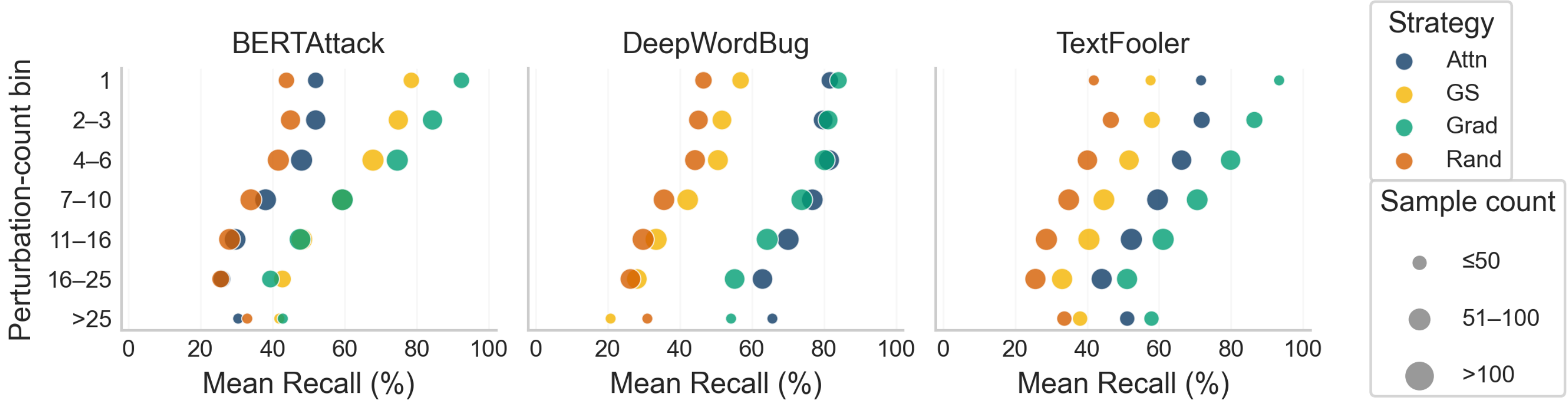}
    \caption{Recall of perturbed words in top-20 rankings by perturbation count bins on Yelp with RoBERTa. Dot size indicates sample count per bin. Higher recall indicates better identification of truly perturbed words. Strategies: Rand (Random), Attn (Attention), GS (Grad-SAM), Grad (Gradient). Similar patterns hold across other dataset-model combinations.}
    \label{fig:binned_performance}
\end{figure*}

\section{Computation Trade-Offs}
\label{sec:comp_tradeoff}

A critical hyperparameter for GPS is determining the optimal number of words $K$ to mask, as too few may miss important adversarial signals, while too many incur unnecessary computational overhead. We evaluate how detection performance varies with $K$, measuring both F1 scores and computational costs to identify the optimal $K$ that balances detection performance with efficiency (Figure~\ref{fig:bertattack_comp}).

GPS achieves remarkable efficiency: with $K{=}5$ capturing over 98\% of the performance observed at $K{=}50$, performance variations beyond this point are negligible ($<0.015$ F1). Computation time scales linearly with $K$, with AG News showing earlier saturation due to shorter document lengths. We find that $K \in [5,10]$ provides an optimal performance-efficiency balance, enabling GPS to operate in resource-constrained environments while maintaining detection quality. The predictable linear scaling allows practitioners to adjust $K$ based on computational budgets without sacrificing reliability, positioning GPS as a practical solution where existing methods may be computationally prohibitive.

\section{Ranking Quality Evaluation}
\label{sec:ndcg_analysis}

Importance heuristics must accurately prioritize words that were actually perturbed during attacks; poor ranking would mean GPS wastes computational resources masking irrelevant words while missing true adversarial modifications. We evaluate each heuristic's effectiveness in ranking perturbed words using Normalized Discounted Cumulative Gain (NDCG)~\cite{Wang2013ATA}, which penalizes relevant items appearing lower in the ranked list (Figure~\ref{fig:ndcg_performance}). This analysis directly tests whether the advantages of gradient-based methods translate to practical perturbation identification.

Using the top-20 candidate words, Gradient Attribution consistently outperforms other heuristics across attack types, showing superior sensitivity to word-level adversarial perturbations. Attention-Rollout performs notably worse against word-level attacks but remains competitive against character-level perturbations. The characteristic spike-dip-recovery pattern in NDCG curves occurs when highly relevant perturbed words appear at top ranks, followed by a drop as irrelevant words enter, then gradual recovery as more perturbed words are found at lower ranks. Gradient-Attribution consistently places perturbed words at the highest ranks. We find these patterns remain consistent across datasets and model variants, with ranking performance mirroring detection results. Effective adversarial detection fundamentally depends on accurate perturbation identification.

\subsection{Robustness to Perturbation Density}
\label{sec:binned_analysis}

Real-world adversarial attacks vary significantly in intensity: some make minimal edits to avoid detection, while others heavily modify text to ensure success (Figure \ref{fig:binned_performance}). Understanding how each heuristic performs across this spectrum is critical for robust detection, as a method that only works on lightly perturbed examples has limited practical value. We sort samples into perturbation-count bins and compute the mean recall of the top-20 candidate words in each bin to quantify how identification quality scales with attack intensity.

Gradient-Attribution exceeds 80\% in the sparse 1-6 range across BERT-Attack, DeepWordBug, and TextFooler. Attention drops sharply once perturbations surpass six words, falling below 40\% for the most heavily perturbed samples; random shows similar decline. This performance gap directly accounts for the weaker detection scores reported in Section \ref{sec:detection_performance}. Identical trends across all three attacks confirm that perturbation density, rather than attack mechanism, drives this failure mode. Gradient-based heuristics maintain higher word-level localization irrespective of perturbation budget, while attention-based methods lose discrimination as adversarial modifications accumulate. This explains why gradient attribution consistently outperforms other approaches across diverse attack scenarios.

\section{Relationship Between Perturbation Identification and Detection Performance}
\label{sec:corr_analysis}

A fundamental question in adversarial detection research is whether methods that excel at identifying specific perturbations necessarily translate to superior detection performance. Understanding this relationship is critical for developing principled approaches to adversarial defense and determining when explanation-based evaluation metrics like NDCG truly reflect detector quality. We investigate whether effective perturbation identification directly correlates with detection accuracy across different attack types and datasets.

\subsection{Dataset Correlations}
We compute Spearman's rank correlation ($\rho$)~\cite{Schober2018CorrelationCoefficients} between detection accuracy and perturbation identification quality (measured by NDCG@20) across all configurations (Table~\ref{tab:global_dataset_corr}). AG News and Yelp show strong positive correlations, establishing that for these datasets, heuristics that better identify perturbations consistently achieve higher detection accuracy. Gradient-based heuristics excel in both perturbation identification and detection under these conditions. IMDB departs from this pattern, showing no significant correlation between perturbation identification and detection performance. The dataset-dependent patterns reveal that the relationship between explanation quality and detection effectiveness is not universal.

\subsection{Attack-Specific Correlation Patterns}

\begin{table}[t]
\centering
\small
\setlength{\tabcolsep}{4pt}
\begin{tabular*}{\columnwidth}{@{\extracolsep{\fill}}lrrrr@{}}
\toprule
\textbf{Dataset} & \textbf{$\rho$} & \textbf{p-value} & \textbf{q-value} & \textbf{n} \\
\midrule
Global & 0.365 & 0.002 & -- & 72 \\
AG News & \textbf{0.903} & $<$0.001 & \textbf{$<$0.001} & 24 \\
Yelp & \textbf{0.723} & $<$0.001 & \textbf{$<$0.001} & 24 \\
IMDB & 0.255 & 0.230 & 0.230 & 24 \\
\bottomrule
\end{tabular*}
\caption{Spearman's correlation ($\rho$) between detection accuracy and NDCG@20. The global correlation across all configurations is moderate, although AG News and Yelp show strong, significant correlations. q-values are Benjamini-Hochberg FDR-corrected~\cite{benjamini1995} with significant values (q $<$ 0.05) in bold.}
\label{tab:global_dataset_corr}
\end{table}

Analyzing correlations by attack type (Table~\ref{tab:attack_corr}) reveals that the relationship between perturbation identification and detection performance depends critically on attack type. Word-level attacks show strong positive correlations between perturbation identification and detection accuracy; when heuristics accurately rank these perturbations, detectors achieve better performance. DeepWordBug presents a fundamentally different pattern, showing no correlation. Character-level attacks operate through different mechanisms where NDCG-based perturbation identification becomes less relevant, and alternative detection mechanisms dominate.

\section{Conclusion}
\label{sec:conclusion}

We introduced Guided Perturbation Sensitivity (GPS), an adversarial text detector that exploits a fundamental property of adversarial examples: their embedding representations are measurably less stable than those of benign text. Adversarial inputs exhibit approximately 2$\times$ higher sensitivity to strategic word masking compared to benign text, a pattern consistent across importance heuristics. By measuring this embedding drift, GPS provides an empirical link between the theoretical instability of adversarial examples near decision boundaries and practical detection in NLP systems.

Our evaluation across 18 configurations reveals that effective adversarial detection is attack-type specific. Word-level attacks exhibit a strong correlation ($\rho > 0.65$) between perturbation identification quality and detection accuracy, validating that gradient-based importance ranking directly enables effective detection. Character-level attacks exhibit no such correlation, operating through different embedding disruption patterns. Cross-architecture transfer experiments further reveal that embedding sensitivity patterns learned on larger models transfer effectively to smaller architectures, indicating that adversarial signatures generalize across model capacities.

GPS achieves 85\%+ detection accuracy while generalizing across datasets, architectures, and attack types without retraining. Its linear scaling with $K$ enables practitioners to balance accuracy against computational cost, with $K{=}5$ capturing 98\% of peak performance. Limitations include the requirement for white-box model access and labeled training data for the BiLSTM detector. Future work should explore adaptive selection of $K$ based on input characteristics and ensemble strategies combining gradient and attention heuristics to capture both word-level and character-level attack signatures. Beyond detection, embedding instability analysis may inform the design of inherently robust architectures and more targeted adversarial training strategies.

\begin{table}[t]
\centering
\small
\setlength{\tabcolsep}{4pt}
\begin{tabular*}{\columnwidth}{@{\extracolsep{\fill}}lrrrr@{}}
\toprule
\textbf{Attack Type} & \textbf{$\rho$} & \textbf{p-value} & \textbf{q-value} & \textbf{n} \\
\midrule
BERT-Attack & \textbf{0.655} & $<$0.001 & \textbf{0.002} & 24 \\
TextFooler & \textbf{0.517} & 0.010 & \textbf{0.015} & 24 \\
DeepWordBug & -0.103 & 0.633 & 0.633 & 24 \\
\bottomrule
\end{tabular*}
\caption{Spearman's correlation ($\rho$) between detection accuracy and NDCG@20 by attack type. Word-level attacks show significant positive correlations, while character-level attacks show slight negative correlation, suggesting different detection mechanisms operate for different attack types.}
\label{tab:attack_corr}
\end{table}

\section*{Acknowledgments}

Research partly supported by NSF grant 2244279, ARO grant W911NF-23-1-0191 and a US Department of Transportation grant for CyberCare. Verma is the founder of Everest Cyber Security and Analytics, Inc.

\bibliography{gps_references}

\appendix

\section{BiLSTM Architecture}
\label{sec:bilstm_arch}

\begin{figure}[htbp] 
  \centering
  \resizebox{\columnwidth}{!}{
  \begin{tikzpicture}[node distance=12mm and 4mm] 

    \node[io_node] (Z) {\faFilter\; \textbf{Input Trace}\\$\mathbf{Z}\!\in\!\mathbb{R}^{B \times N\times2}$};
    \node[preprocess_node, below=of Z] (aug)
          {\faCogs\; \textbf{Augment Features}\\(+ Mask \& Pos. Encoding)\\$\mathbf{X}\!\in\!\mathbb{R}^{B \times N\times4}$};

    \node[projection_node, below=of aug] (proj)
          {\faBolt\; \textbf{Input Projection}\\(Linear(4,64) $\rightarrow$ LayerNorm $\rightarrow$ GELU)};
    \node[core_lstm_node, below=of proj] (bilstm)
          {\faStream\; \textbf{2-Layer BiLSTM}\\(64 hidden/direction, Dropout 0.3)};

    \node[attention_node, below=of bilstm, yshift=-14mm] (attention) {\faProjectDiagram\; Multi-Head Attention\\(2 Heads, Masked)};
    \node[max_pool_node, left=2mm of attention] (max_pool) {\faCompress\; Max Pooling};
    \node[avg_pool_node, right=2mm of attention] (avg_pool) {\faEquals\; Average Pooling\\(Masked)};
    
    \node[combine_node, below=of attention, yshift=-14mm] (concat) {\faLayerGroup\; \textbf{Feature Concatenation}\\(Outputs from Max, Attention, Avg)};
    
    \node[classifier_node, below=of concat] (head) 
          {\faCheckCircle\; \textbf{Classification Head}\\(FC $\rightarrow$ GELU $\rightarrow$ Dropout $\rightarrow$ FC $\rightarrow$ Logits)};

    \draw[conn] (Z.south) -- (aug.north);
    \draw[conn] (aug.south) -- node[dimtag_right, xshift=-2mm, pos=0.5] {$B \!\times\! N \!\times\! 4$} (proj.north);
    \draw[conn] (proj.south) -- node[dimtag_right, xshift=-2mm, pos=0.5] {$B \!\times\! N \!\times\! 64$} (bilstm.north);
    
    \coordinate (branch_hub) at ($(bilstm.south)!0.5!(attention.north)$); 
    \draw[conn] (bilstm.south) -- node[dimtag_right, xshift=-2mm, pos=0.5] {$B \!\times\! N \!\times\! 128$} (branch_hub);
    \draw[conn] (branch_hub) -- (max_pool.north);
    \draw[conn] (branch_hub) -- (attention.north);
    \draw[conn] (branch_hub) -- (avg_pool.north);

    \coordinate (merge_hub) at ($(attention.south)!0.5!(concat.north)$); 
    \draw[conn] (max_pool.south) -- (merge_hub) node[pos=0.5, dimtag_right,xshift=-5mm, yshift=1mm] {$B \!\times\! 128$};
    \draw[conn] (attention.south) -- (merge_hub) node[pos=0.5, dimtag_right, yshift=1mm] {$B \!\times\! 128$};
    \draw[conn] (avg_pool.south) -- (merge_hub) node[pos=0.5, dimtag_right, xshift=5mm, yshift=1mm] {$B \!\times\! 128$};
    \draw[conn] (merge_hub) -- node[dimtag_right, xshift=-2mm, pos=0.5] {$B \!\times\! 384$} (concat.north);
    
    \draw[conn] (concat.south) -- (head.north);
    
    \begin{pgfonlayer}{background}
        \node[stageplate=paultolgreen_plate, fit=(Z) (aug), inner sep=4mm] {};
        \node[stageplate=paultolblue_plate, fit=(proj) (bilstm), inner sep=4mm] {};
        \node[stageplate=paultolorange_plate, fit=(max_pool) (attention) (avg_pool) (branch_hub) (merge_hub), inner sep=4mm] {};
        \node[stageplate=paultolpurple_plate, fit=(concat), inner sep=4mm] {};
        \node[stageplate=paultolred_plate, fit=(head), inner sep=4mm] {};
    \end{pgfonlayer}{background}
  \end{tikzpicture}%
  } 
    \caption{\textbf{Architecture of the BiLSTM‑based adversarial detector.}
    The input trace $\mathbf{Z}$ is augmented with a binary mask identifying non‑zero positions and a linear positional encoding, then normalized to form $\mathbf{X}\in\mathbb{R}^{N\times C}$.
    After an input projection, $\mathbf{X}$ passes through a 2‑layer Bidirectional LSTM.
    Sequence outputs are summarized by a 2‑head self‑attention block, max‑pooling, and mean‑pooling; the three resulting vectors are concatenated.
    A feed‑forward classification head maps the pooled representation to logits for the \emph{benign} vs.\ \emph{adversarial} classes.}
    \label{fig:bilstm-architecture} 
\end{figure}
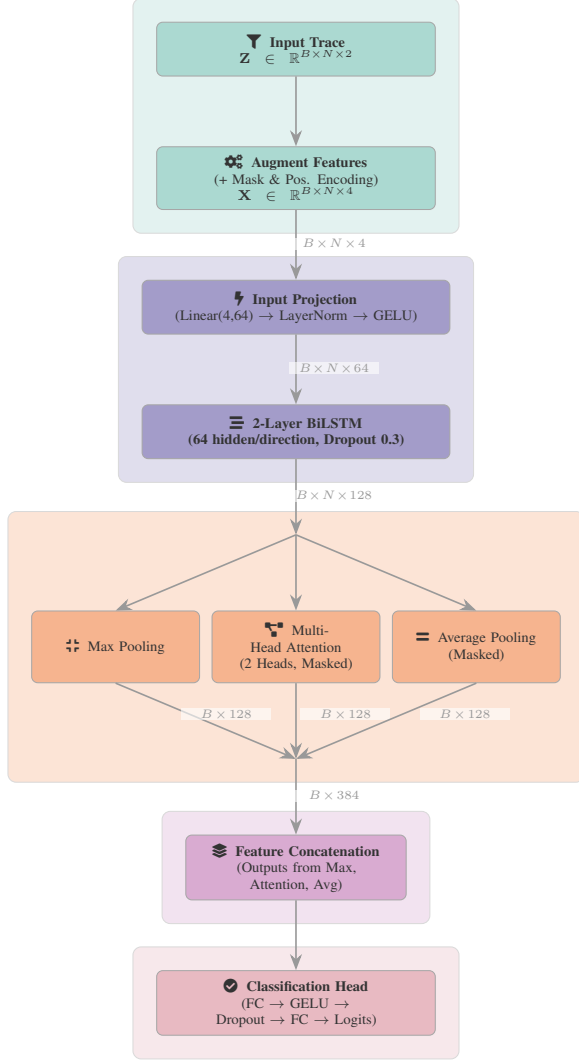

Our detector processes the sensitivity-importance traces using a carefully designed BiLSTM architecture that captures sequential patterns in adversarial examples. Figure~\ref{fig:bilstm-architecture} illustrates the complete network architecture.

\paragraph{Input Processing.}
Before feeding the trace $\mathbf{Z}$ into the LSTM layers, we normalize the sensitivity and importance channels based on the non-zero values observed during training. We augment the input features by incorporating a binary mask channel (indicating non-zero entries, primarily for handling variable lengths and zero-padding implicitly) and a linear positional encoding channel, normalized to the range [0, 1]. This results in an input tensor $\mathbf{X} \in \mathbb{R}^{N \times C}$, where $C$ includes the original sensitivity/importance channels plus the added mask and positional channels.

\paragraph{Core Architecture.}
The input tensor $\mathbf{X}$ is first passed through an input projection layer (a linear layer followed by Layer Normalization~\cite{ba2016layernormalization} and GELU activation~\cite{hendrycks2023gaussianerrorlinearunits}) to map the features into the model's hidden dimension space. The core of the detector consists of a 2-layer BiLSTM with a hidden dimension size of 64 per direction. The bidirectional nature allows the model to process the sequence, leveraging both past and future context at each position. Dropout (rate 0.3) is applied between LSTM layers for regularization.

\paragraph{Pooling and Attention.}
To aggregate information across the sequence dimension, we employ a combination of pooling strategies. The output sequence from the BiLSTM is processed by:
\begin{enumerate}
    \item A multi-head attention mechanism (2 heads) to compute a context vector that adaptively weights different sequence positions based on their relevance. Masking is applied during attention calculation to ignore padding or zero-valued positions.
    \item Max pooling across the sequence dimension to capture the most salient features.
    \item Average pooling across the sequence dimension (masked to ignore padding) to capture overall sequence characteristics.
\end{enumerate}
The resulting vectors from attention, max pooling, and average pooling are concatenated.

\paragraph{Classification Head.}
The concatenated pooled representation is passed through a final classification head consisting of a fully connected layer with GELU activation, followed by dropout (rate 0.3), and a final linear layer producing logits for the two classes (benign/adversarial).

\paragraph{Training.}
The model is trained using the AdamW optimizer with a learning rate of 0.0005. We use a batch size of 32 and employ early stopping based on validation performance (F1-score) with a patience of 5 epochs, training for a maximum of 40 epochs.

\section{Extended Performance Analysis}
\label{sec:full_perf}

This section provides additional performance metrics and analyses that complement the main paper results, offering insights into our RS framework's characteristics and trade-offs.

\subsection{Accuracy-Efficiency Trade-offs}
\label{sec:comp_time}

\begin{figure*}[htbp]
    \centering
    \includegraphics[width=\textwidth]{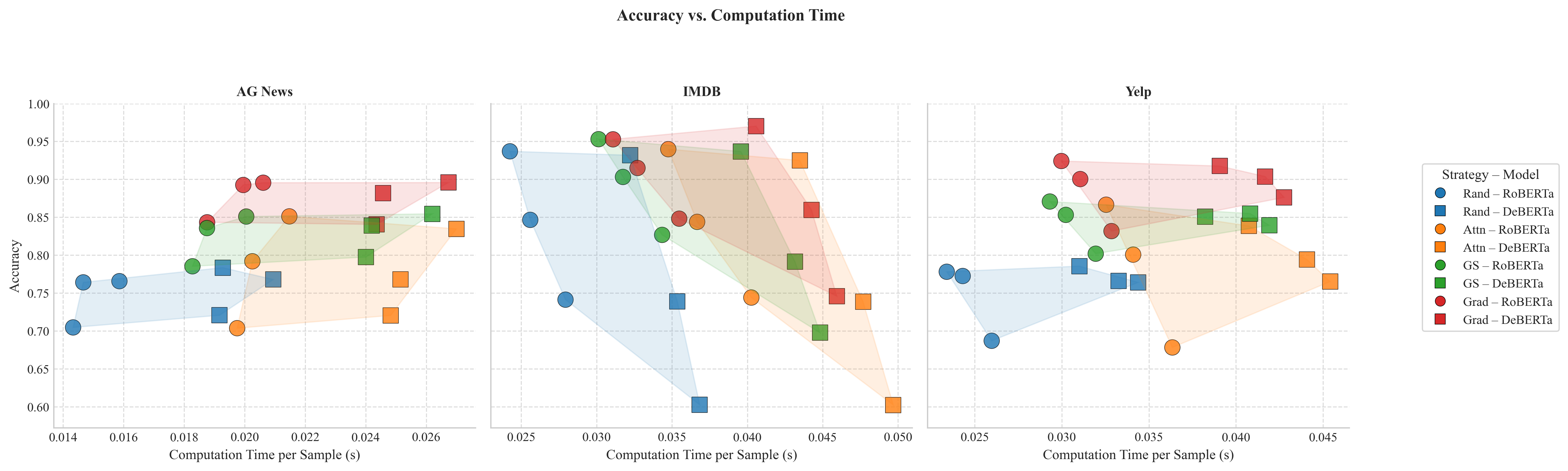} 
    \caption{Detection accuracy versus computation time per sample (seconds) for generating the sensitivity-importance trace using different heuristics. Each point represents a specific combination of heuristic, victim model (RoBERTa: circles, DeBERTa: squares), dataset, and attack type. Shaded areas represent convex hulls for each strategy across models within a dataset. Gradient-based methods (Grad, Grad-SAM) cluster towards higher accuracy and higher computation time, while Attention and Random are faster but less accurate.}
    \label{fig:ex_accuracy_vs_time}
\end{figure*}

Our analysis of computational efficiency reveals a pattern: all importance heuristics (Gradient, Grad-SAM, and Attention) demonstrate remarkably similar computation times per sample, typically within 0.001-0.005 seconds of each other (Figure~\ref{fig:ex_accuracy_vs_time}). This contradicts the conventional expectation that gradient-based methods would incur substantially higher computational costs due to their backpropagation requirements. The attention rollout extraction process involves operations with comparable complexity, including averaging attention weights across heads, computing attention rollout through matrix multiplications, and processing multi-layer attention patterns.

The clear efficiency distinction emerges when comparing our RS framework against TextShield, which requires 5-10× more computation time while often achieving lower accuracy. Random word selection provides a modest efficiency advantage (approximately 25-30\% faster than other heuristics) but at a significant performance cost, particularly for word-level attacks. Within architecture comparisons, DeBERTa (184M parameters) consistently exhibits higher computation times than RoBERTa (125M parameters) across all heuristics, reflecting its larger capacity.

This analysis presents an advantageous scenario for practitioners: gradient-based methods deliver superior detection accuracy without the expected computational penalty, making them the preferred choice for most deployments. The GPU-accelerated computation of gradients in modern deep learning frameworks effectively mitigates the complexity difference between gradient and attention-based approaches. These findings highlight that the accuracy-efficiency trade-off in adversarial detection depends more on model architecture and detector design than on the choice between gradient and attention-based importance heuristics.

\subsection{Integrated Gradients: Steps vs. Performance}
\label{int_grad}

\begin{table*}[htbp]       
  \centering
  \resizebox{\textwidth}{!}{%
    \begin{tabular}{@{}lccccccc@{}}
      \toprule
      \textbf{Steps} &
      \textbf{Accuracy} &
      \textbf{Precision} &
      \textbf{Recall} &
      \textbf{F1} &
      \textbf{AUC} &
      \textbf{nDCG} &
      \textbf{Time (s/sample)} \\
      \midrule
      10  & 0.7978 & 0.7849 & 0.8244 & 0.8028 & 0.8780 & 0.6400 & 0.0426 \\
      25  & 0.8120 & \textbf{0.8094} & 0.8168 & 0.8128 & 0.8856 & 0.6435 & 0.0843 \\
      50  & 0.8110 & 0.7742 & 0.8824 & 0.8236 & 0.9045 & \textbf{0.6449} & 0.1551 \\
      100 & \textbf{0.8228} & 0.7716 & \textbf{0.9200} & \textbf{0.8387} & \textbf{0.9152} & 0.6446 & 0.2956 \\
      \bottomrule
    \end{tabular}
  }
  \caption{Integrated Gradients detector results  
           (victim: \textbf{RoBERTa}, attack: \textbf{TextFooler}, dataset: \textbf{AG News}).
           }
  \label{tab:ig_roberta_textfooler_agnews}
\end{table*}

We also investigated using Integrated Gradients (IG)~\cite{integrated_grad} as an alternative importance heuristic. Table~\ref{tab:ig_roberta_textfooler_agnews} shows how the number of integration steps affects detector performance and computation time.

Our analysis of Integrated Gradients reveals a nuanced relationship between integration steps and detection performance. As integration steps increase from 10 to 100, F1 score improves from 0.803 to 0.839, with the most significant gains occurring in the 25-50 step range. This improvement comes with an expected computational cost; processing time scales linearly with step count, increasing from 0.043s to 0.296s per sample. Notably, perturbation identification quality (measured by nDCG) plateaus after just 25 steps (0.644→0.645), despite continued improvements in detection metrics at higher step counts. We also observe that the precision-recall balance shifts toward higher recall with more integration steps, demonstrating the detector becomes more robust at identifying adversarial samples, albeit at the cost of slightly more false positives. This analysis demonstrates that while basic gradient-based approaches offer a good balance of performance and efficiency for most applications, Integrated Gradients with 50-100 steps can provide improved detection capabilities in scenarios where computational resources permit the additional processing time.

\section{Extended Correlation Analysis}
\label{sec:ex_corr}

\begin{figure*}[htbp]
    \centering
    \includegraphics[width=\textwidth]{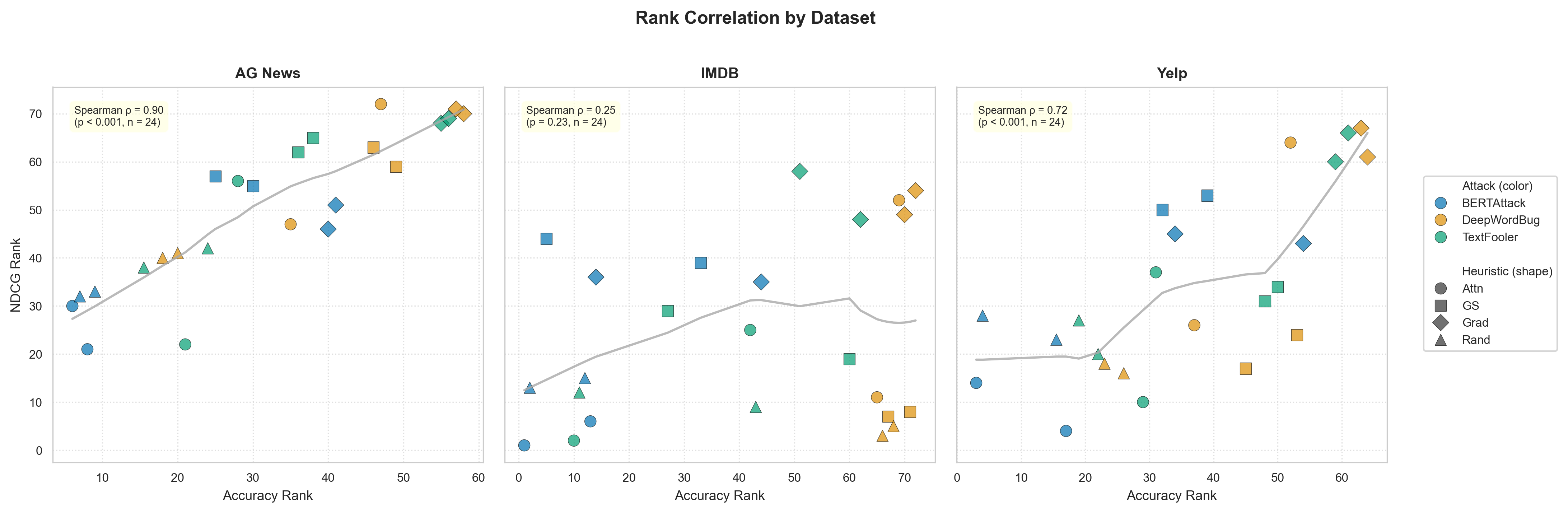} 
    \caption{Dataset-specific rank correlations between detector accuracy and explanation NDCG. Subplots for AG News, IMDB, and Yelp present individual experimental configurations (n=24 each), with attack type encoded by color and explanation heuristic by shape. LOWESS smoothing curves (grey lines) and Spearman's $\rho$ statistics are shown. AG News ($\rho=0.90, p<0.001$) and Yelp ($\rho=0.72, p<0.001$) exhibit strong positive correlations. Conversely, IMDB ($\rho=0.25, p=0.23$) shows no significant correlation, highlighting the dataset-dependent nature of this relationship. The distribution of attack-heuristic combinations suggests their varied influence on both metrics within and across datasets.}
    \label{fig:dataset_scatters}
\end{figure*}

Table~\ref{tab:ex_dataset_backbone_corr} examines whether model architecture affects these correlations. Interestingly, RoBERTa and DeBERTa show nearly identical correlation patterns within each dataset, suggesting the relationship between perturbation identification and detection performance is primarily determined by dataset characteristics rather than backbone architecture. To further explore the dataset-specific relationships, Figure~\ref{fig:dataset_scatters} presents scatter plots of accuracy ranks versus NDCG ranks for each dataset, with attack types encoded by color and explanation heuristics by shape.

The dataset-specific correlation analysis reveals striking differences in how perturbation identification quality relates to detection performance. AG News exhibits an exceptionally strong correlation ($\rho$=0.90) with data points closely following a diagonal pattern, demonstrating that perturbation identification quality directly translates to detection performance for this dataset. We find that the clustering of points remains consistent across attack methods. Yelp similarly maintains a strong positive correlation ($\rho$=0.72), though with greater variance. The predominantly diagonal LOWESS curve confirms that better perturbation identification generally leads to improved detection performance on this dataset, despite some outliers.

In sharp contrast, IMDB shows no significant correlation ($\rho$=0.25, p=0.23) between NDCG and accuracy ranks. The widely dispersed data points and relatively flat LOWESS curve suggest that for IMDB, factors beyond perturbation identification quality, possibly related to the dataset's longer text length or greater semantic complexity, play a more dominant role in determining detection performance. Together, these analyses reveal that the relationship between perturbation identification and detection is both attack-dependent and dataset-dependent, highlighting the complex nature of adversarial text detection.

\begin{table}[htbp]
\centering
\small
\setlength{\tabcolsep}{4pt}
\begin{tabular*}{\columnwidth}{@{\extracolsep{\fill}}lrrrr@{}}
\toprule
\textbf{Dataset-Model} & \textbf{$\rho$} & \textbf{p-value} & \textbf{q-value} & \textbf{n} \\
\midrule
AG News-RoBERTa & \textbf{0.930} & $<$0.001 & \textbf{$<$0.001} & 12 \\
AG News-DeBERTa & \textbf{0.916} & $<$0.001 & \textbf{$<$0.001} & 12 \\
Yelp-RoBERTa & \textbf{0.727} & 0.007 & \textbf{0.011} & 12 \\
Yelp-DeBERTa & \textbf{0.755} & 0.005 & \textbf{0.009} & 12 \\
IMDB-RoBERTa & 0.231 & 0.471 & 0.471 & 12 \\
IMDB-DeBERTa & 0.231 & 0.471 & 0.471 & 12 \\
\bottomrule
\end{tabular*}
\caption{Spearman's correlation ($\rho$) between detection accuracy and NDCG@20 by dataset and backbone model. The relationship patterns are consistent across backbones within each dataset, suggesting dataset characteristics rather than model architecture determine correlation strength.}
\label{tab:ex_dataset_backbone_corr}
\end{table}

\section{Extended Perturbation Identification}
\label{sec:ex_pert_id}

This section extends our perturbation identification analysis to all dataset-model combinations, providing a comprehensive view of how different importance heuristics perform across datasets and backbone architectures.

\subsection{Ranking Quality}
\label{sec:ex_rank_qual}

\begin{figure*}[htbp]
    \centering
    
    \begin{minipage}{\textwidth}
        \centering
        \includegraphics[width=\textwidth]{figs/ndcg_at_k_Yelp_RoBERTa.png}
        \captionof{figure}{NDCG@k performance on Yelp with RoBERTa across three attack types. This is the same figure shown in the main text for reference.}
        \label{fig:ex_yelp_ndcg_roberta}
    \end{minipage}
    
    \vspace{0.5cm}
    
    \begin{minipage}{\textwidth}
        \centering
        \includegraphics[width=\textwidth]{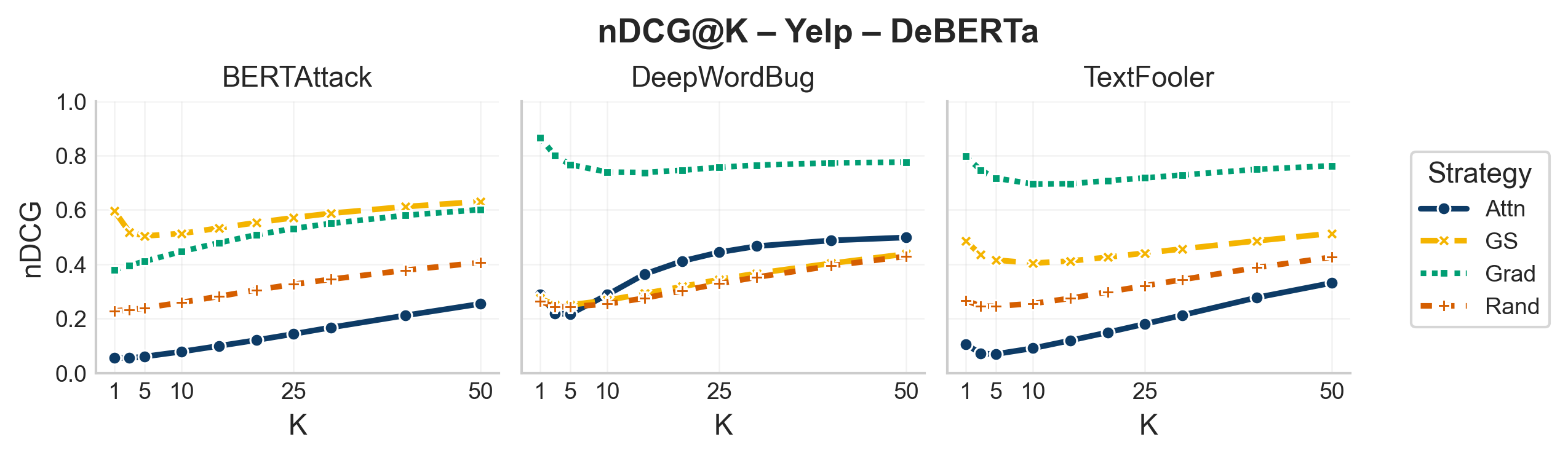}
        \captionof{figure}{NDCG@k performance on Yelp with DeBERTa across three attack types. The pattern is consistent with RoBERTa, showing gradient-based methods are superior at ranking perturbed words regardless of backbone architecture.}
        \label{fig:ex_yelp_ndcg_deberta}
    \end{minipage}
\end{figure*}

\begin{figure*}[htbp]
    \centering
    
    \begin{minipage}{\textwidth}
        \centering
        \includegraphics[width=\textwidth]{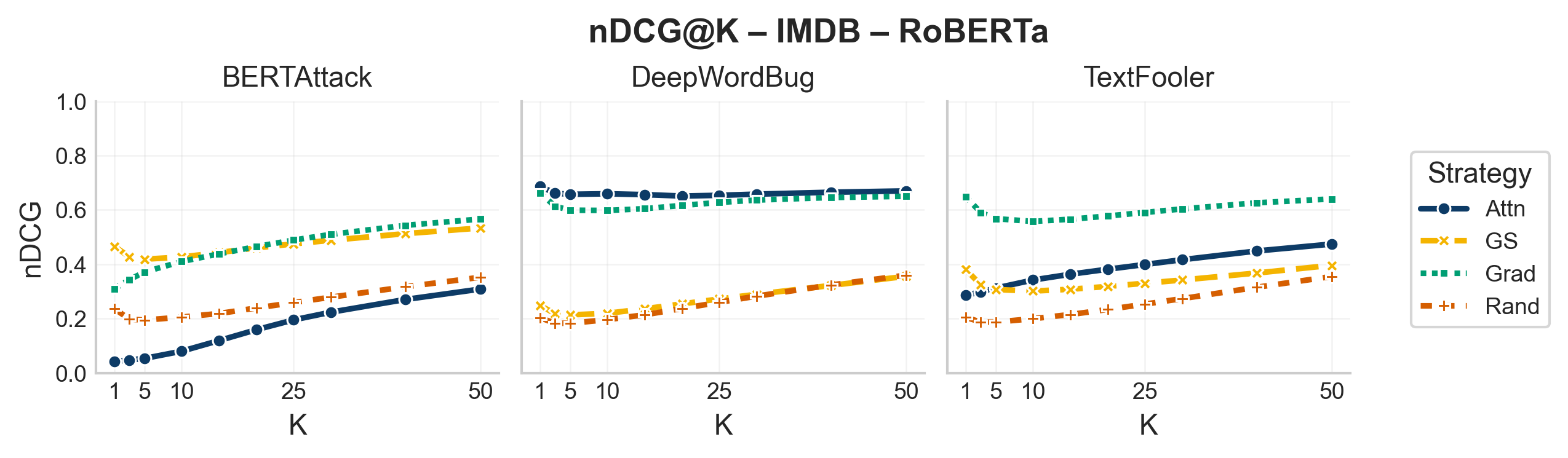}
        \captionof{figure}{NDCG@k performance on IMDB with RoBERTa across three attack types. The longer text length in IMDB leads to lower overall NDCG scores across all heuristics, but gradient-based methods still maintain their ranking advantage.}
        \label{fig:ex_imdb_ndcg_roberta}
    \end{minipage}
    
    \vspace{0.5cm}
    
    \begin{minipage}{\textwidth}
        \centering
        \includegraphics[width=\textwidth]{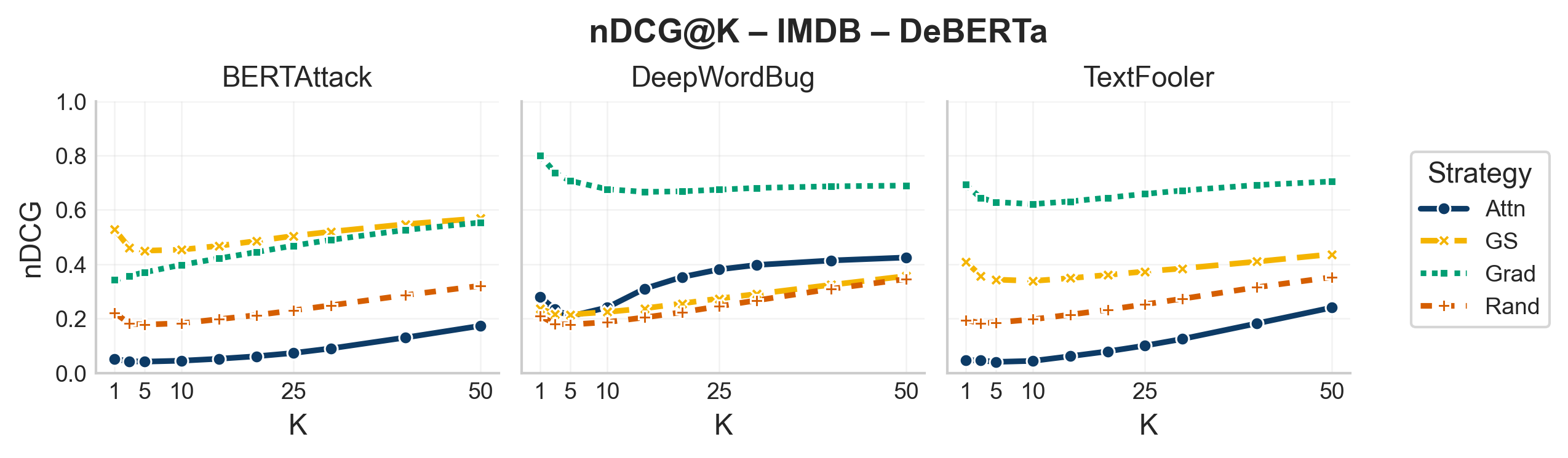}
        \captionof{figure}{NDCG@k performance on IMDB with DeBERTa across three attack types. The performance gap between gradient-based methods and attention is particularly pronounced for word-level attacks.}
        \label{fig:ex_imdb_ndcg_deberta}
    \end{minipage}
\end{figure*}

\begin{figure*}[htbp]
    \centering
    
    \begin{minipage}{\textwidth}
        \centering
        \includegraphics[width=\textwidth]{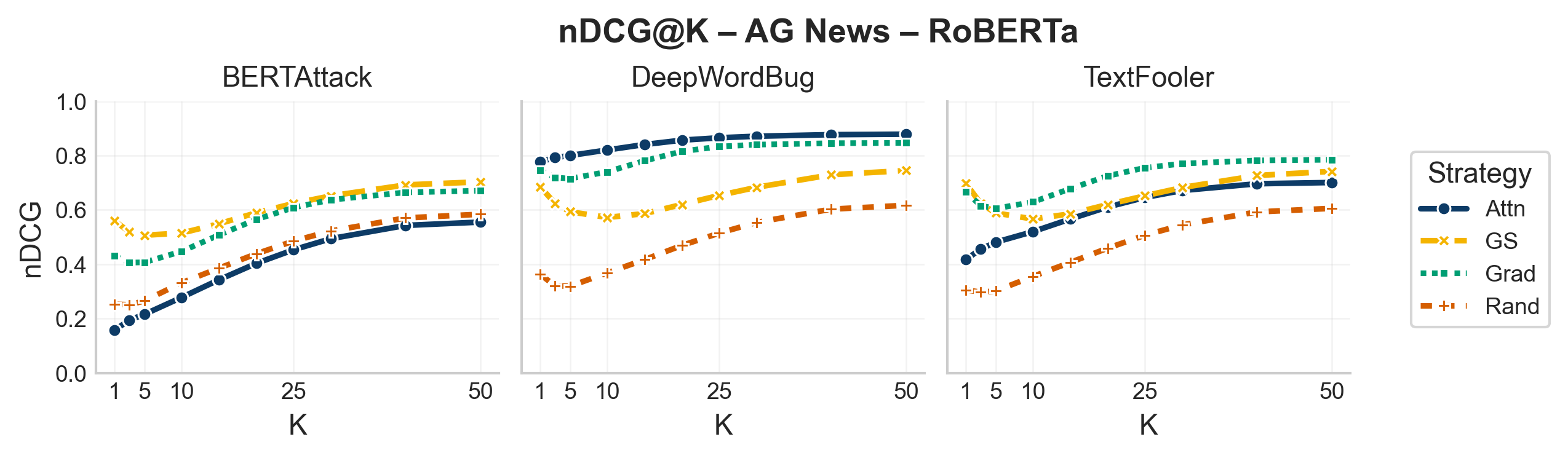}
        \captionof{figure}{NDCG@k performance on AG News with RoBERTa across three attack types. Topic classification data shows particularly strong performance from gradient-based methods, with Grad achieving NDCG@20 values above 0.7 for TextFooler.}
        \label{fig:ex_ag_ndcg_roberta}
    \end{minipage}
    
    \vspace{0.5cm}
    
    \begin{minipage}{\textwidth}
        \centering
        \includegraphics[width=\textwidth]{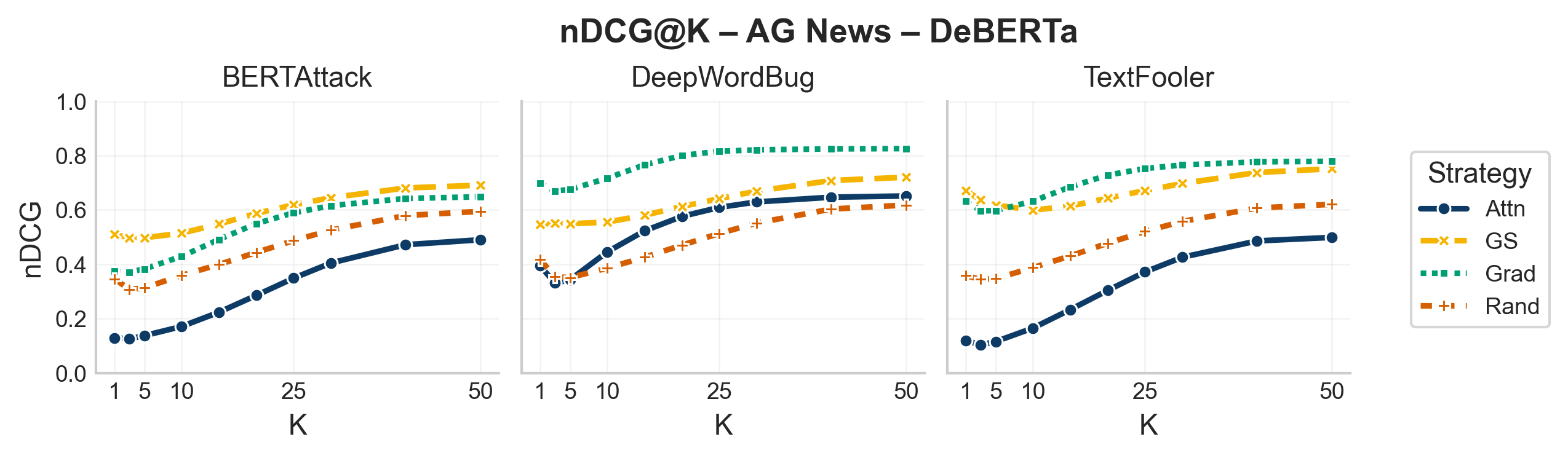}
        \captionof{figure}{NDCG@k performance on AG News with DeBERTa across three attack types. The trend mirrors RoBERTa, with topic classification texts showing strong differentiation between importance heuristics.}
        \label{fig:ex_ag_ndcg_deberta}
    \end{minipage}
\end{figure*}

In figures \ref{fig:ex_yelp_ndcg_roberta} through \ref{fig:ex_ag_ndcg_deberta}, across all datasets and models, we observe that gradient-based methods (Grad and Grad-SAM) consistently outperform attention-based and random baselines at ranking perturbed words, particularly for word-level attacks (TextFooler and BERT-Attack). For character-level attacks (DeepWordBug), attention sometimes approaches gradient-based performance, especially in shorter texts. Additionally, the topical nature of AG News appears to make perturbation identification more straightforward compared to sentiment-based datasets.

\subsection{Perturbation Density}
\label{sec:ex_pert_den}

\begin{figure*}[htbp]
    \centering
    
    \begin{minipage}{\textwidth}
        \centering
        \includegraphics[width=\textwidth]{figs/bin_dotplot_cal_Yelp_RoBERTa.png}
        \captionof{figure}{Mean recall across perturbation count bins on Yelp with RoBERTa. This is the same figure shown in the main text for reference.}
        \label{fig:ex_binned_yelp_roberta}
    \end{minipage}
    
    \vspace{0.5cm}
    
    \begin{minipage}{\textwidth}
        \centering
        \includegraphics[width=\textwidth]{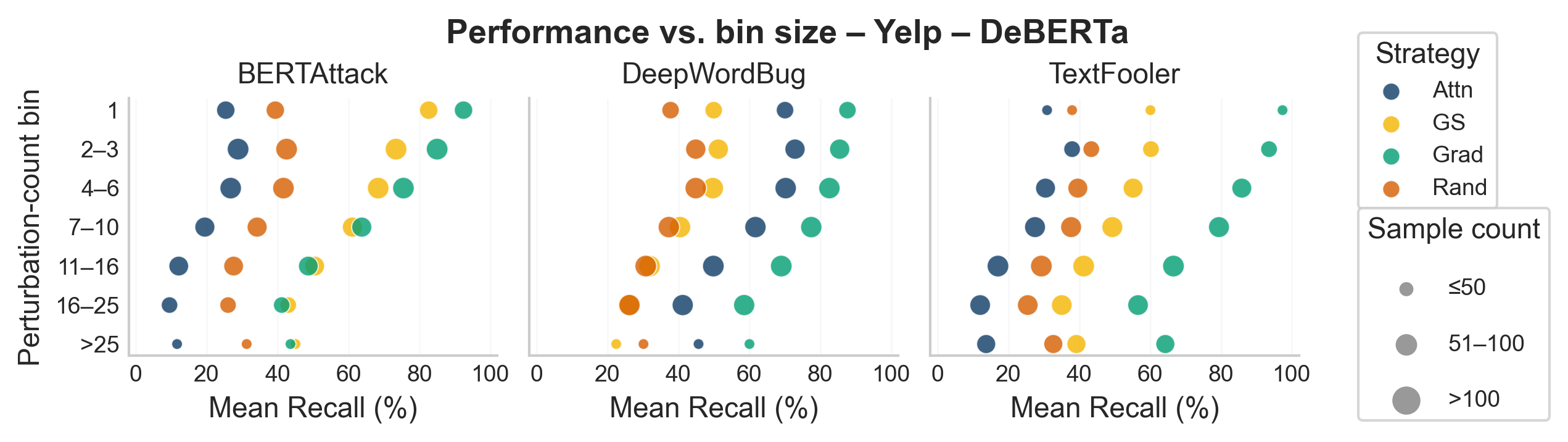}
        \captionof{figure}{Mean recall across perturbation count bins on Yelp with DeBERTa. The trend follows RoBERTa, with gradient-based methods showing substantially better robustness to higher perturbation counts.}
        \label{fig:ex_binned_yelp_deberta}
    \end{minipage}
\end{figure*}

\begin{figure*}[htbp]
    \centering
    
    \begin{minipage}{\textwidth}
        \centering
        \includegraphics[width=\textwidth]{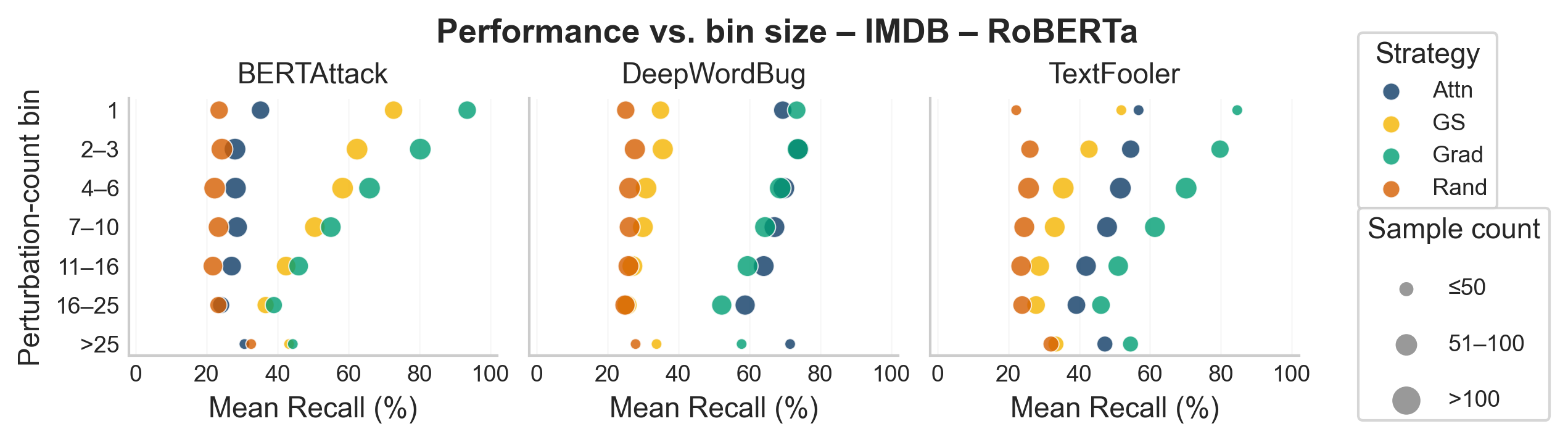}
        \captionof{figure}{Mean recall across perturbation count bins on IMDB with RoBERTa. The longer sequences in IMDB present a greater challenge, with all methods showing lower recall for heavily perturbed examples.}
        \label{fig:ex_binned_imdb_roberta}
    \end{minipage}
    
    \vspace{0.5cm}
    
    \begin{minipage}{\textwidth}
        \centering
        \includegraphics[width=\textwidth]{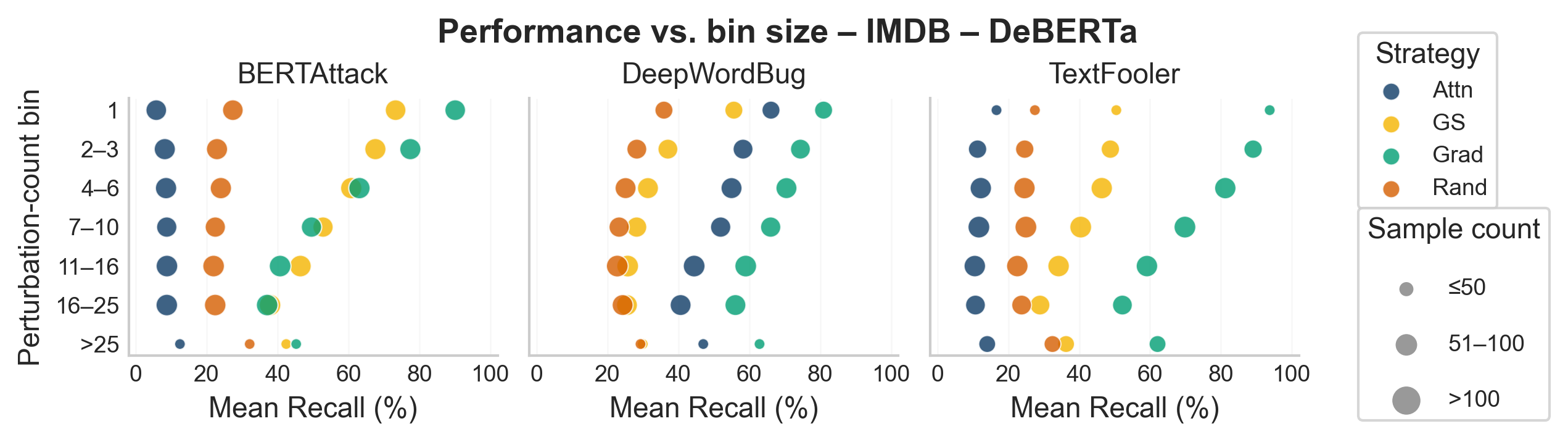}
        \captionof{figure}{Mean recall across perturbation count bins on IMDB with DeBERTa. The increased sequence length in IMDB makes perturbation identification more challenging, but gradient-based methods still maintain their advantage.}
        \label{fig:ex_binned_imdb_deberta}
    \end{minipage}
\end{figure*}

\begin{figure*}[htbp]
    \centering
    
    \begin{minipage}{\textwidth}
        \centering
        \includegraphics[width=\textwidth]{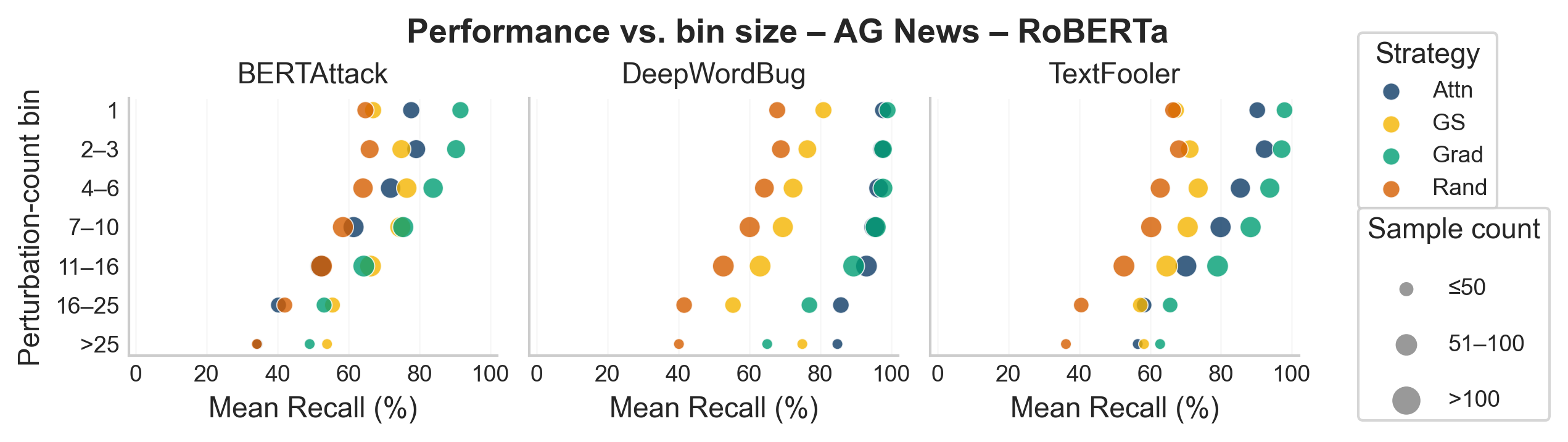}
        \captionof{figure}{Mean recall across perturbation count bins on AG News with RoBERTa. In this topic classification dataset, gradient-based methods show remarkable robustness against perturbation density increases.}
        \label{fig:ex_binned_ag_roberta}
    \end{minipage}
    
    \vspace{0.5cm}
    
    \begin{minipage}{\textwidth}
        \centering
        \includegraphics[width=\textwidth]{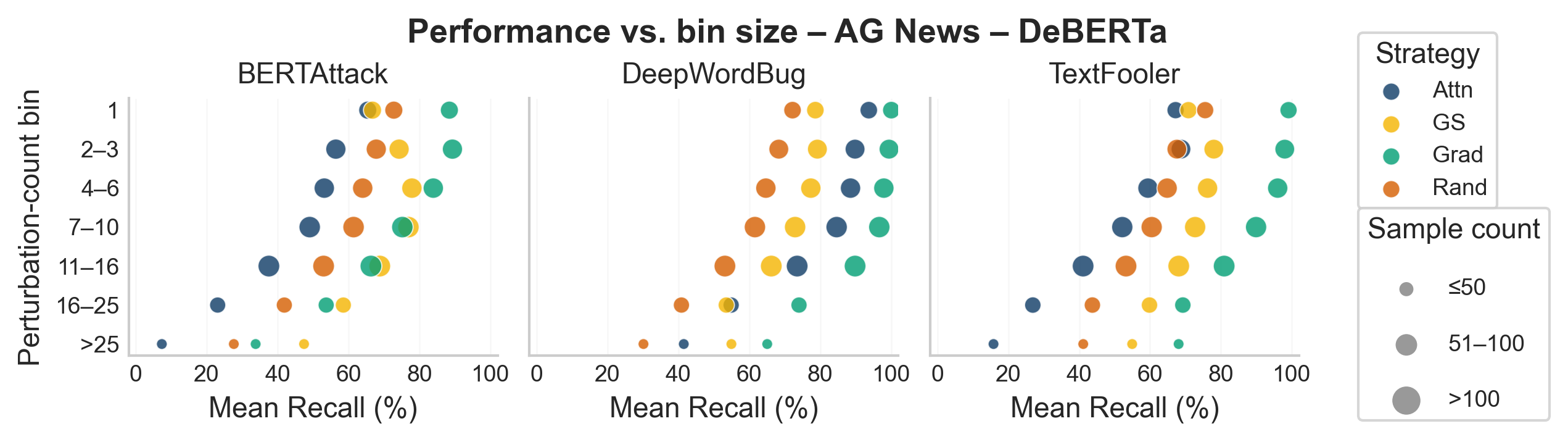}
        \captionof{figure}{Mean recall across perturbation count bins on AG News with DeBERTa. The consistency of patterns between RoBERTa and DeBERTa confirms the stability of our findings across model architectures.}
        \label{fig:ex_binned_ag_deberta}
    \end{minipage}
\end{figure*}

In figures \ref{fig:ex_binned_yelp_roberta} through \ref{fig:ex_binned_ag_deberta}, our binned recall analysis reveals that all heuristics tend to degrade as perturbation counts increase, but at significantly different rates. Gradient-based methods maintain relatively high recall even for heavily perturbed examples, while attention-based approaches show a steeper decline, particularly for word-level attacks. This trend holds across datasets and models, with some dataset-specific variations in overall recall levels, likely attributable to differences in text length and semantic complexity. AG News shows the most stable recall across perturbation densities, while IMDB (with its longer text length) presents a greater challenge, especially at higher perturbation counts. 



\end{document}